\documentclass[sigconf,nonacm]{acmart}

\AtBeginDocument{%
  \providecommand\BibTeX{{%
    \normalfont B\kern-0.5em{\scshape i\kern-0.25em b}\kern-0.8em\TeX}}}

\copyrightyear{2023}
\acmYear{2023}
\acmDOI{10.NNNN/NNNNNNN.NNNNNNN}

\acmPrice{15.00}
\acmISBN{978-1-4503-XXXX-X/18/06}
\usepackage{tikz}

\newcommand{\etal}{{\em et al.}\xspace}
\usepackage{xspace}
\usepackage{setspace}
\newcommand{\note}[1]{}

\usepackage[linesnumbered,ruled]{algorithm2e}
\SetKw{KwBy}{by}

\newcommand{\ie}{{\em i.e.,}\xspace}

\usepackage{amsfonts,chemarrow,balance}
\usepackage{amsmath} 
\usepackage{graphicx}
\usepackage{url}
\usepackage{mathenv}
\usepackage{color}
\usepackage{breqn}
 \usepackage{algpseudocode,caption}
\newcommand{\BfPara}[1]{{\noindent\bf#1.}\xspace}
\usepackage{multirow}
\usepackage{float}
\usepackage{threeparttable}
\usepackage[inline]{enumitem}
\usepackage{booktabs}
\usepackage{makecell}
\usepackage{subcaption}

\newcommand{\ecir}[0]{\tikz\draw[black,fill=white] (0,0) circle (.6ex);}
\newcommand{\fcir}[0]{\tikz\draw[black,fill=black] (0,0) circle (.6ex);}

\begin{document}

\title{Burning the Adversarial Bridges: Robust Windows Malware Detection Against Binary-level Mutations}

\author{Ahmed Abusnaina}
\affiliation{%
  \institution{Meta Inc.}
  \city{Seattle, WA}
  \country{United States}}
\email{abusnaina@meta.com}

\author{Yizhen Wang}
\affiliation{%
  \institution{Visa Research}
  \city{Palo Alto, CA}
  \country{United States}}
\email{yizhewan@visa.com}

\author{Sunpreet Arora}
\affiliation{%
  \institution{Visa Research}
  \city{Palo Alto, CA}
  \country{United States}}
\email{sunarora@visa.com}

\author{Ke Wang}
\affiliation{%
  \institution{Visa Research}
  \city{Palo Alto, CA}
  \country{United States}}
\email{kewang@visa.com}

\author{Mihai Christodorescu}
\affiliation{%
  \institution{Visa Research}
  \city{Palo Alto, CA}
  \country{United States}}
\email{mihai.christodorescu@visa.com}

\author{David Mohaisen}
\affiliation{%
  \institution{University of Central Florida}
  \city{Orlando, FL}
  \country{United States}}
\email{david.mohaisen@ucf.edu}

\renewcommand{\shortauthors}{Abusnaina, et al.}

\begin{abstract}
  The continuous malware development cycle, by adding malicious capabilities and obfuscating existing features, hinders the efforts of malware detection engines in timely detecting new threats. However, it is argued that malware evolvement, in general, requires re-compiling and re-distributing the malicious software, slowing the spread of malware mutations. In this work, we provide a counterargument to this assertion. We leverage the concepts from the binary analysis domain, e.g., binary manipulation, to demonstrate their effectiveness in evading state-of-the-art detection engines. In particular, we evaluate the robustness of four state-of-the-art malware detection models, alongside five industry-standard malware detection engines. Our findings highlight that, among others, these models are susceptible to various binary manipulation-based threats, including binary padding and section injection attacks. 
  Toward robust malware detection, we explore the attack surface of existing malware detection systems. We conduct root-cause analyses of the practical binary-level black-box adversarial malware examples. Additionally, we uncover the sensitivity of volatile features within the detection engines and exhibit their exploitability.
  
  Highlighting volatile information channels within the software, we introduce three software pre-processing steps to eliminate the attack surface, namely, padding removal, software stripping, and inter-section information resetting. 
  Further, to counter the emerging section injection attacks, we propose a graph-based section-dependent information extraction scheme for software representation. 
  The proposed scheme leverages aggregated information within various sections in the software to enable robust malware detection and mitigate adversarial settings. 
  Our experimental results show that traditional malware detection models are ineffective against adversarial threats. 
  However, the attack surface can be largely reduced by eliminating the volatile information. Therefore, we propose simple-yet-effective methods to mitigate the impacts of binary manipulation attacks. Overall, our graph-based malware detection scheme can accurately detect malware with an area under the curve score of 88.32\% and a score of 88.19\% under a combination of binary manipulation attacks, exhibiting the efficiency of our proposed scheme. 
\end{abstract}



\maketitle
\vspace{-2mm}
\section{Introduction}
The rapid growth in the malware has pushed the detection frameworks a step behind in the arms race for a timely malware detection and identification.
This is mainly attributed to the heightened volume of malware observed on a daily basis~\cite{VirusTotalStats}.
Traditionally, detection frameworks have employed signature- and heuristics-based approaches for malware detection. However, these techniques' performance, while largely efficient across malicious software with similar characteristics, decreases significantly when encountering a new malware with unique capabilities. 
This motivated a broad spectrum of malware detection using machine learning techniques, capable of detecting a wide range of malware properties~\cite{AlrawiLVSMAM21,RaffSN17}.

Machine learning-based techniques are, to some extent, effective in detecting malware and generalizing to new unseen malicious patterns. However, the performance of such techniques is bounded by the notion of {\em ``concept drift''}~\cite{jordaney2017transcend}.
The fundamental idea behind concept drift is to invoke a rapid malware mutation, causing a continuous shift in the feature space, to eventually cause misclassification.
To combat this critical issue, a commonly used approach is model retraining by incorporating new malware samples in the training process~\cite{KantchelianAHIMTGJT2013}.
However, this approach is rendered infeasible due to the large amount of reported new malware that emerge daily, an approximate of 1.5 million new malware per day~\cite{VirusTotalStats}. 
While some of these malware samples belong to the newly emerging malware category (\ie family), most of them have resulted from either (i) code-level mutations or (ii) binary-level mutations. 

The code-level mutation is the process in which malware code is modified to add, remove, or modify functionalities within the malware.
Binary-level mutation, on the other hand, is the process where the software is modified after compilation, on the binary level. 
This results in more restrictions in comparison with the code-level mutations. For instance, removing contents may render the malware unexecutable, and information adding or modification is limited to designated locations. 
While, At first glance, code-level mutations allow for stronger adversarial capabilities, they require direct and mostly manual interference from the malware author. The interaction between the malware author and the source code requires re-compiling and re-distributing the malicious software.
On the other hand, the process of binary-level mutations can be automated, and done on the end-point devices, without the interaction of the malware author, nor re-compiling the source code of the malware~\cite{castro2019armed}. 

Bounded by its limitations, binary-level mutations, where adversaries can only add new information and behaviors to the malware, can be decoded using advanced dynamic analysis and behavioral fingerprinting techniques. Such techniques can uncover the malicious functionality, mitigating the effect of the added information. However, it is argued that these techniques are resource and time-consuming~\cite{ZhangQW20,CaiMRY18}, and can take minutes or even hours per malware sample. 
The dynamic analysis becomes infeasible in light of the ongoing growth in malicious software numbers. With the rapid increase of emerging malware, binary-level mutations worsen the already-existing bottleneck dependency on malware analysis frameworks. 
To reduce the dependency on such frameworks, detection frameworks deployed lightweight static-based machine learning malware detection solutions on the end-point devices (\ie personal computers). 
However, recent studies have shown their ineffectiveness against binary-level malware mutations~\cite{DCBLAR2020,DemetrioBLRA21}.

Addressing this concern, Incer~\cite{IncerTA018} proposed the usage of hard to manipulate monotonically increasing features for malware detection. Monotonically increasing features ensure that the addition of information, resulting from the binary-level manipulations, does not cause misclassification. However, this comes at the cost of a high reduction in malware detection performance. 
In this work, we revisit this challenge. First, we consider a threat scenario where the malware is continuously mutating while spreading among the end-point devices \textbf{(see \autoref{sec:threat})}. Then, we investigate the existing lightweight binary-level malware detection techniques to understand their capabilities and limitations. Afterward, we characterize the commonly-made mistakes while implementing lightweight malware detection frameworks, providing guidelines for robust malware detection. Finally, inspired by the previous point, we introduce a component-based robust malware detection system with a reduced malware performance reduction trade-off.

\begin{figure}[t]
\centering
\includegraphics[width=0.48\textwidth]{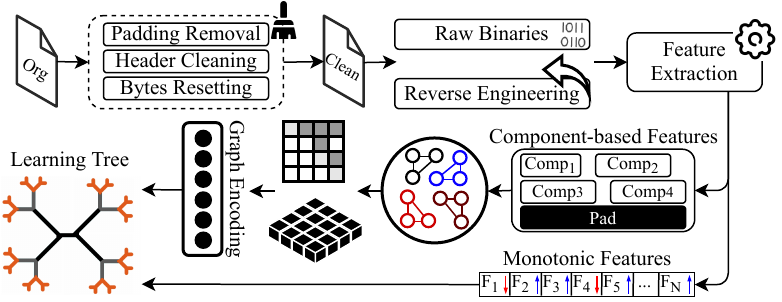}\vspace{-3mm}
\caption{The overview of the proposed system design. The design consists of four modules, including software pre-processing, feature extraction, component-based graph encoding, and malware detection. }
\label{fig:SystemDesign}\vspace{-5mm}
\end{figure}

\autoref{fig:SystemDesign} shows our proposed graph-based system design for binary-level manipulations-robust malware detection. The design consists of four main components: (i) Software pre-processing component for volatile information channels removal, including padding removal, header and debugging information removal, and inter-section unmapped bytes resetting. (ii) Feature extraction, where various per-component (\ie software section) raw binary- and reverse engineering-level features are extracted from the software. (iii) Component-based graph encoding, in which, the different representations of each component interact, using information aggregation, generating robust software representation. (iv) Decision tree-based malware detection model for robust malware identification. Through experimental evaluation, we show that the proposed system design achieves a robust malware detection performance of up to 88.19\%. 

\BfPara{Contributions} Our contributions are as follows:

\begin{itemize}
    \item \textbf{\underline{Malware Threat Surface:}} We analyze the malware threat surface, and the root causes of the effectiveness of various binary-level manipulations, including binary padding, and header information manipulation. Our analysis highlighted that the issue lines within the selected information channels for software feature representation, and can be largely addressed by revisiting the used features.
    \item \textbf{\underline{Software Volatile Features Removal:}} Inline with the previous contribution, we proposed several software binary-level pre-processing techniques for volatile (\ie exploitable) information channels removal from the feature representation of the software. Our findings support that most binary-level manipulations can be mitigated following easy-to-apply modifications to the software.
    \item \textbf{\underline{Graph Encoding-based Malware Detection:}} We propose a graph-based representation for robust malware detection against the state-of-the-art component injection attacks (\ie section injection). The experimental results show that such attacks can be mitigated with a low detection performance trade-off. 
\end{itemize}


\section{Background}\label{sec:background}
In the section, we provide a brief background on malware detection and adversarial settings in the context of malware detection domain.

\subsection{Malware Detection}
The prior work in this space explores the potential of machine learning algorithms for building effective malware detection systems~\cite{AlasmaryAPCNM19,AnwarAPW20,vasan2020image,mercaldo2020deep}. 
The performance of such systems largely depends on the choice of representations, generated by static and dynamic analysis techniques. 

With advances in learning theory, the application of learning algorithms to defend systems against malware attacks has provided remarkable success. In order to apply advances in learning theory, malware binaries are transformed into different representations that are machine learning-compatible~\cite{ZhangQYOH16}. For example, Cui \etal~\cite{CuiXCCWC18} introduced a malware detection method using deep learning by transforming the malicious code into grayscale images, achieving an accuracy rate of 94.5\% on the Vision Research Lab dataset~\cite{nataraj2011malware}, and showing better performance than both static~\cite{kang2015detecting} and dynamic~\cite{Al-DujailiHHO18} feature-based representations. Similarly, Ni \etal~\cite{NiQZ18} proposed a malware detection system trained over 10,805 grayscale images associated with nine different Windows malware families with comparable success. Fu \etal~\cite{FuXWLS18}, on the other hand, proposed a malware detector using an RF model trained over colored images generated from malware binaries, achieving a detection accuracy of 97.47\% and a family classification accuracy of 96.85\%.

Moreover, Pajouh \etal\cite{haddadpajouh2018deep} detected ARM-based malware targeting IoT devices using an LSTM-based architecture to model opcode sequences, achieving detection accuracy of 98.18\%. Furthermore, Wang \etal~\cite{WangGZOXLG17} proposed an adversary-aware neural network technique for malware detection by leveraging feature nullification. 
Furthermore, Anderson~\etal~\cite{anderson2018ember} proposed a feature representation leveraging a variety of statically extracted information, including strings, function calls, and used libraries, among others, to effectively detect malware with high efficiency.

The aforementioned literature aimed to identify malicious behavior using various representations. In this work, we leverage several aforementioned representations and techniques for malware detection tasks, evaluating the robustness of the said approaches.

\subsection{Adversarial Machine Learning}
Machine/deep learning networks are widely used in security-related tasks, including malware detection~\cite{MohaisenAM15, AlasmaryKAPCAANM19}. However, it has been shown that deep learning-based models are vulnerable against adversarial attacks~\cite{miyatoMKNI15}. Unfortunately, such a behavior can be a critical issue in malware detection systems, where misclassifying malware as benign may result in compromising the underlying service~\cite{AbusnainaAASNM19,AbusnainaMYM19}.

Various adversarial machine learning attack methods in the context of image classification have been introduced to generate AEs~\cite{Carlini017,Moosavi-Dezfooli16}. While initially targetting image classification frameworks, recent studies investigated generating AEs in the context of malware detection~\cite{SuciuCJ19}. For instance, Grosse~\etal~\cite{GrossePMBM17} implemented an augmented adversarial crafting algorithm to generate AEs, misleading a CNN-based classifier to misclassify 63\% of the malware samples to benign.

The detection of the AEs is challenging~\cite{Carlini017_2}. While work on detecting AEs in the context of IoT malware detection is very limited, multiple studies attempt to detect them in the context of image classification~\cite{Xu0Q18,LiL17,MetzenGFB17}, achieving detection accuracy of 20\% to 90\%.
In this work, we implement various volatile information elimination approaches, alongside with graph encoding-based malware detection framework for adversarial malware mitigation, achieving a state-of-the-art robust malware detection performance of up to 88.19\% with adversarial mitigation capabilities.

\begin{figure}[t]
\centering
\includegraphics[width=0.26\textwidth]{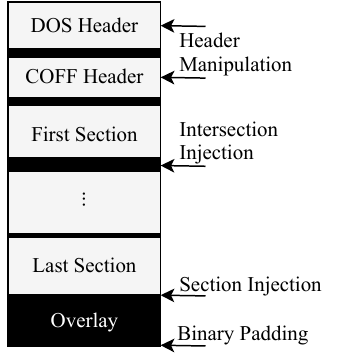}\vspace{-3mm}
\caption{The Windows file format. Different binary-level manipulation and injection attacks take place to cause misclassification. The existing attack channels are mainly attributed to the ability to add and modify a large portion of the software binaries.}
\label{fig:attackChannels}\vspace{-5mm}
\end{figure}

\section{Threat Model}\label{sec:threat}
The literature on machine learning-based malware detection has shown the vulnerability of such techniques against binary-level mutations. 
Such vulnerabilities rendered the models useless, as they (i) are easy to generate, (ii) do not require compiling the source code, and (iii) can be done automatically before re-spreading the malware from one endpoint to another. 
In the following, we discuss the binary-level adversarial capabilities, including insights regarding their effectiveness, shown in~\autoref{fig:attackChannels}.

\subsection{Adversarial Capabilities}\label{sec:adversarial_capabilities}
In these settings, the adversary applies perturbation directly to the binary sequences of the malicious file. The perturbation can be either (1) modifying the values of existing bytes, or (2) adding new byte sequences to the file. Such perturbation should be carefully applied to ensure that the (i) functionality, (ii) executability, and (iii) malicious behavior of the software are intact.  
In the following, we investigate the binary-level malware mutations that cause effective malware misclassification. 

\BfPara{Header Information Manipulation~\cite{DCBLAR2020}}
In this attack, the adversary modifies the arbitrary values within the program header or DOS header of the program to cause misclassification. Information includes, but is not limited to, size of image, program signature, characteristics flags, time-date stamp, signature, and the number of sections. Such information is easy to modify, as they do not contribute to the functionality of the program

\noindent\textit{\underline{Insights on Effectiveness:}} This attack is effective when the malware detection frameworks utilize easy-to-modify (\ie volatile) header information for malware detection~\cite{anderson2018ember}. While such information may momentarily increase the malware detection performance, as malware within the same category and time period may have shared header information, it might be exploited, via removal or modification, to reduce the detection performance or confidence. 

\BfPara{Binary Padding~\cite{KolosnjajiDBMGE18}}
In this attack, the adversary appends binaries at the end of the program binaries. This process is functionality-preserving and widely used for misclassifying raw binaries-based classifiers. The padded binaries can be generated using different approaches, including benign injection, random injection, or gradient padding. 

\noindent\textit{\underline{Insights on Effectiveness:}} While padded bytes are not mapped to any functionality, nor scanned by the operating system on the run-time, this approach take advantage of the on-the-fly lightweight raw binary-level malware detection approaches, including bytes sequences~\cite{RaffBSBCN18}, software visualization~\cite{SuVPSFS18,makandar2017malware}, and bytes n-grams and histograms~\cite{yousefi2018malytics,barr2021survivalism}. Such techniques, while being lightweight and utilized at the end-point devices, can be exploited to cause misclassification using binary padding.

\BfPara{Inter-section Injection~\cite{DCBLAR2020,SuciuCJ19}}
In contrast to the binary padding, this approach does not increase the size of the program, but rather manipulates the ``unused'' (\ie unmapped) bytes of the program to cause misclassification. This, however, limit the space of perturbation and may result in a reduced attack surface. The unused bytes are defined as bytes resulting from the memory page allocation process, where bytes are padded between sections due to the difference between the virtual size and allocated memory size.

\noindent\textit{\underline{Insights on Effectiveness:}} Similar to binary padding, this attack exploits the lack of software structural analysis in raw binary-level detection models. However, it is harder to detect, as it does not increase the size of the file, in contrast to the binary padding.

\BfPara{Section Injection~\cite{DemetrioBLRA21}}
The software consists of the header, containing general information regarding the executable, and multiple sections, each of which contains information regarding the functionality or data of the malware. In this attack, the adversary introduces a new section that is mapped in the file header, and may be executable (\ie ``.text'' backdoor injection).

\noindent\textit{\underline{Insights on Effectiveness:}} This is considered stronger than the previous adversarial settings, as the added section appears in the program analysis, reverse engineering of the software, and is mapped to the allocated location. In addition, new functionalities and behavior may be added to the preserved malicious functionality. Mitigating such an attack is challenging, as identifying injected sections requires deep analysis of the control and data flow, or dynamic analysis of the malicious behavior, a solution that is time and resource consuming, and not feasible at the end-point devices.

\begin{figure}[t]
\centering
\includegraphics[width=0.45\textwidth]{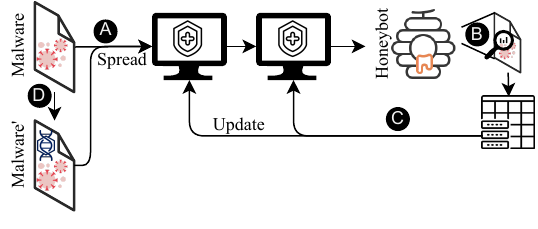}\vspace{-6mm}
\caption{Code level perturbation. (A) The malware author develops a new malware that is undetected by the existing malware detection frameworks. (B) The malware is spread between vulnerable devices, and eventually captured by a honeypot for behavioral analysis. (C) Once analyzed, the malware characteristics are used to update lightweight malware defense on end-point devices. (D) Malware authors apply code-level mutations to surpass the updated detection frameworks.}
\label{fig:CodeLevelPert}\vspace{-3mm}
\end{figure}

\subsection{Malware Mutation Scenarios}
The literature on robust malware detection focused on the code-level mutations of malware, where the malicious code is assumed to be retrievable, and re-compilable on end devices. While this is true for Android malware, Windows malware mutations are time-consuming and mainly manual. In this work, we introduce binary-level robust malware detection against binary-level malware mutations. In the following, we discuss both mutation scenarios, and the limitations of each of them.

\BfPara{Code-level Mutations: Attack Scenario} Figure~\ref{fig:CodeLevelPert} shows the attack scenario of the code-level mutations. First, the malware author develops unseen (\ie undetected) malware that spreads among the exploitable end-point devices over time. The infected devices have a lightweight malware detection model that is operated by a malware detection service provider. Initially, the malware will be undetected on such devices. However, after a period of time, a malware honeypot, configured by the service provider, captures the malware and forwards it for further analysis. Upon deep dynamic analysis, several characteristics of the malware are extracted and used to update the end-point devices defense systems to accurately detect the malware. 

To avoid future detection, the malware author mutates the malware by adding or modifying the functionality of the malware, starting a new cycle of malware re-spreading. These settings require a direct and most likely manual interaction of the malware author. 

\begin{figure}[t]
\centering
\includegraphics[width=0.45\textwidth]{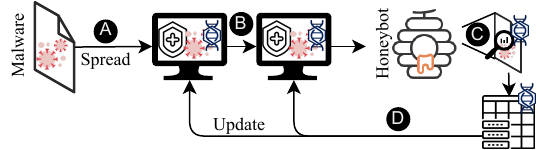}\vspace{-3mm}
\caption{The threat of automated binary-level mutation. (A) Malware author develops a malware new to existing malware detectors. (B) The malware mutates when spreading between vulnerable devices. (C) Only a fraction of the mutants are captured by honeypots and analyzed to update the detector. (D) The lightweight malware defenses installed on end-point devices are updated based on the mutants rather than the original malware, causing failures in detection as the malware continuously mutates.}
\label{fig:BinaryLevelPert}\vspace{-3mm}
\end{figure}

\BfPara{Binary-level Mutations: Attack Scenario} Figure~\ref{fig:BinaryLevelPert} shows the attack scenario of the binary-level mutations. First, similar to the previous setting, the malware author develops unseen malware that spread among the exploitable end-point devices. However, before infecting a new device, the malware mutates using functionality preserving binary-level mutation techniques. While this is not feasible in the previous settings, as the source code and compilation capabilities should be present to mutate the malware, binary-level mutations are automated with low computational requirements. Over time, the malware will be captured by the honeypot after $n$ mutations, and the end-point devices will be updated using the mutated malware information, rather than the original malware information. This is problematic as (i) malware authors can enforce fake patterns in their binary-level mutations, and (ii) the updated defense systems only accommodate for one mutation, which is, in this scenario, different from the mutations that exist on the infected devices.

The latter scenario motivates for binary mutations-robust malware detection frameworks. In this work, we introduce guidelines to effectively build robust malware detection models against existing binary-level manipulations, with a low trade-off on the baseline detection performance. We limit our adversarial settings to the ones that can be conducted on the end-point device, leveraging binary-level black-box lightweight attacks that can effectively mislead the detection engines.

\subsection{Black-box Adversarial Attacks}\label{sec:adversarial_attacks}
In order to evaluate the robustness of various machine learning-based malware detection models, we implemented simple binary-level black-box attacks that can be launched on the endpoint device without the need to access the underlying detection model or its parameters. We argue that white-box access to the machine learning configurations is unrealistic in the context of malware detection, and operating under black-box settings highlights the vulnerability of these models in practice. The following is a brief description of the leveraged black-box attacks used in this work.

\BfPara{Header Information Stripping}
Static analysis-based malware detection models, in general, leverage easy to modify header information to detect malware samples. Eliminating the header information disrupt the extraction of such patterns, causing misclassification. In this work, we also consider removing the debugging information, as it may contain patterns that are recognizable by the detection engine.

\BfPara{Benign Binaries Padding}
Several pieces of literature have focused on binary padding as a simple form of adversarial examples. In this work, we follow the benign injection method, where benign binary sequences are padded to the malicious software after EOF (end of file). For the generalizability of this approach, the selected bytes sequences are among the operating system files (\ie within \textit{``Windows/System32''} directory). 

\BfPara{Inter-section Injection}
The values of the unmapped bytes between the sections of the PE file are either randomized or compiler dependent. In the latter case, the detection engines may fingerprint the patterns within these areas, enabling accurate detection of malware. In this attack, we randomize the value of the unmapped bytes within different sections of the program, erasing the potential recognizable patterns.

\BfPara{Benign Section Injection}
In this attack, instead of injecting benign binaries to the end of the malicious software, selected sections within the benign software are injected in the malicious samples using section injection attack. Similar to benign binaries padding, the selected sections belong to benign files among the operating system files (\ie within \textit{``Windows/System32''} directory).

\section{Dataset \& Experimental Setup}\label{sec:Dataset}

\subsection{Dataset Overview}\label{sec:Dataset Overview}
In this work, we obtained 9,000 malicious Windows software binaries from VirusShare~\cite{VirusShare}. The dataset includes malicious samples with first seen dates from 2017 to 2020. Moreover, we compiled a benign dataset of 8,817 executable binaries obtained from default configurations of operating system instances of Windows XP, 7, 8, 8.1, and 10. As the default configuration of each operating system only contains software provisioned by Microsoft, we assume that all those software binaries are benign, by default, which is confirmed through our sanity check.

\BfPara{Selection of Malware Samples} In this work, we consider a malware dataset that is diverse. Per section~\ref{sec:adversarial_capabilities}, an adversary can slightly modify the malware to avoid the classification, while preserving the main functionality. In this work, we filtered out all malicious samples with the same assigned family and EMBER~\cite{anderson2018ember} original or monotonic~\cite{IncerTA018} representation. If two malicious samples are assigned the same family and share the same feature representation, only the earliest is considered in our dataset. This ensures the removal of a high number of ``malicious duplicates'' from our compiled dataset. This process, in turn, will result in reduced malware detection baseline accuracy, as the malicious dataset is more diverse, hardening the extraction of useful malicious patterns.

\BfPara{Ground Truth Class} Next, we used \textit{VirusTotal}~\cite{VirusTotal} to examine the malicious and benign samples in our dataset to validate the correctness of the labels. Additionally, we obtained the \textit{first seen} date of the malicious samples, which we consider as the ground-truth appearance date in this work. All benign samples were not detected by \textit{VirusTotal}, supporting our default benign assumption.

\BfPara{Adversarial Variants} Leveraging the aforementioned adversarial attacks (refer to section~\ref{sec:adversarial_attacks}), we generate four adversarial variants of each malicious software for bypassing the malware detection frameworks.

\subsection{Software Representation}
Provided that this work is targeting malware detection on the end-point device, we only leverage binary-level efficient approaches for malware detection. Each software, and its different variants, are converted into different feature representations, including:

\begin{enumerate}
    \item \BfPara{Software Visualization~\cite{vasan2020image,mercaldo2020deep}} The grayscale image representation is a common technique for transforming malware samples into a 2D matrix of values, and each of those values is between 0 and 255. Particularly, the byte-code is visualized as a grayscale image of a fixed size of ($h\times w$), where every byte is a pixel in the image.
    \item \BfPara{MalConv~\cite{RaffBSBCN18}} MalConv is a deep neural network trained on raw bytes for malware binary detection instead of deep features, allowing for reduced feature extraction time in comparison to other approaches.
    \item \BfPara{EMBER Representation~\cite{anderson2018ember}} EMBER representation utilizes different software features, including \textit{Hexdump}, \textit{Function Calls} and \textit{Entropies}, \textit{Program Sections Information}, \textit{Relocations}, and \textit{Strings} to convert a software into a vector of size $1\times 2,137$. 
    \item \BfPara{EMBER Monotonic Representation~\cite{IncerTA018}} Monotonic EMBER representation utilizes the monotonically increasing subset of EMBER features for malicious patterns recognition. This requires the removal of features that can decrease in value with the addition of information, such as \textit{Entropy}.  
\end{enumerate}

\subsection{Experiment Setup and Metrics}\label{sec:ExperimentalSetup}
For our experiments, we train machine and deep learning baseline models on 6,000 benign samples, and malware samples that appeared between 2017 and 2019 (6,000 malicious samples), followed by an evaluation of the trained model to detect the malicious samples that appeared in 2020 (a total of 3,000 malicious samples), alongside the remaining 2,817 benign samples. We note that the data splitting takes into consideration balanced training and testing sets for the training process.

\BfPara{Learning Algorithm Selection} In the following, we describe our learning algorithm selection for each software representation approach.

\begin{itemize}
    \item \BfPara{Software Visualization~\cite{vasan2020image,mercaldo2020deep}} We leverage ResNet-110 for malware detection task, feeding the generated images to the CNN-based model.
    \item \BfPara{MalConv~\cite{RaffBSBCN18}} We followed the same deep learning architecture proposed by the authors for MalConv feature representation.
    \item \BfPara{Ember Features-based Model~\cite{anderson2018ember}} We utilized the features extracted by Anderson~\etal for implementing the Ember baseline. Ember features are considered the baseline for malware detection.
    \item \BfPara{Monotonically Increasing Ember model~\cite{IncerTA018}} We follow the implementation in Incer~\etal~\cite{IncerTA018} to implement verifiable robust malware detection based on monotonically increasing features. 
\end{itemize}

\BfPara{Evaluation Metric} The trained models were evaluated using the true positive rate, \ie malware detection rate. The {\em true positive rate} is the ratio between the correctly classified malware and the total number of malware samples. We fixed the false positive rate, defined as the percentage of benign samples incorrectly classified as malware, to 1\% and 3\% in our evaluation. We also report the area under the curve of the receiver operating characteristic (ROC-AUC) in the evaluation to show the classification task potential of the trained models.

\BfPara{Online Detection Engines} Alongside the four aforementioned baseline models, we incorporated the results of different lightweight online detection engines. Namely, we report the results of \textit{``Avast''}, \textit{``Arcabit''}, \textit{``McAfee''}, \textit{``BitDefender''}, and \textit{``Malwarebytes''} detection engines. In the evaluation, the aforementioned engines (in randomized order) are referred to as \textit{``Online Engine --- 1''} to \textit{``Online Engine --- 5''}. This step is necessary to prevent direct comparison between the detection engines, in compliance with the terms and conditions of use.

\section{Robust Malware Detection}\label{sec:robust_detection}
In this section, we provide insights toward robust malware detection. In particular, (i) we first provide a software pre-processing approach to significantly reduce the attack surface of the binary-level manipulations, limiting the adversary capabilities to only section injection (\autoref{sec:preprocessing}). Then, (ii) we formulate the robustness requirements to mitigate the section injection attacks (\autoref{sec:robust_req}). Finally, (iii) we provide a hands-on example of a robust malware detection system with reduced malware detection trade-off (\autoref{sec:sysDesignRobust}).

\subsection{Robust Software Pre-processing}\label{sec:preprocessing}
In this section, we discuss simple yet effective software pre-processing techniques to significantly mitigate the effect of binary-level mutations. This includes (i) software stripping, (ii) excessive length removal, and (iii) bytes resetting. 

\begin{enumerate}
    \item \BfPara{Software Stripping} The process of removing information from executable binaries that is not essential or required for normal and correct execution is known as software stripping. While such information may increase the performance of the detection framework, it is considered a fertile ground for adversary exploitation, particularly the header and debugging information.
    \item \BfPara{Excessive Length Removal}
    We refer to the process of removing the padded binaries that occur after the end of the file, and do not appear in the header or sections information (\ie not mapped to the software) as binary unpadding. Obtaining the information regarding the sections boundaries and virtual sizes can be effectively utilized to omit the padded binaries from the end of the software. 
    \item \BfPara{Bytes Resetting}
    In the software, there exists a space between the mapped sections, mainly caused by the memory paging system and the difference between virtual size and raw size of the section. This space is typically exploited by modifying the byte sequences to generate code caves and different malware mutations. Resetting these bytes to a pre-defined op-code, such as $0x00$, removes the possibility of conducting such adversarial attacks.

\end{enumerate}

The aforementioned pre-processing techniques remove the potential perturbation added by the adversary, except for section injection attack. In the next section, we propose graph-encoding component-based malware detection framework toward robust malware detection under section injection attacks.

\begin{algorithm}[t]
\SetAlgoLined
\KwIn{Detector $D$, Input Software $S$, Graph Attention Network $GAT$}
\KwOut{Detection Label $l$}
$\triangleright$ Pre-process the software to remove attack channels\\
 $S^{'1}$ $\leftarrow$ Strip($S$)\;
 $S^{'2}$ $\leftarrow$ PadRemove($S^{'1}$)\;
 $S^{'3}$ $\leftarrow$ ResetBytes($S^{'2}$)\;

 $\triangleright$ Extract the monotonically increasing features\\

 Mono $\leftarrow$ Extract\_Mono\_Features($S^{'3}$)\;

  $\triangleright$ Extract the component-based features\\
 Sections $\leftarrow$ Sec\_Extract($S^{'3}$)\;
 
  INIT list $C$, graph $G$\;

  \For{Sec $\in$ Sections}{
    INIT subgraph $G_{sub}$\;
    
    $\triangleright$ Extract section related features\\
    Comp\_Rep $\leftarrow$ Extract\_Features(Sec)\;
    $\triangleright$ Build fully connected subgraph\\
    \For{Rep $\in$ Comps\_Reps}{
        $G_{sub}$.add\_node(Rep)
    }
    $G_{sub}$.fully\_connected()\;
    
    $\triangleright$ Add subgraph to graph space\\
    $G$.add($G_{sub}$)\;
  }

 $C$  $\leftarrow$ $GAT$.encode($G$)\;

 INIT list $F$\;
 $F$.append(Mono)\;
 $F$.append($C$)\;
 
 $l$ $\leftarrow$ D.predict($F$)\;
 Return $l$\;
 
 \caption{The proposed graph-based encoding pipeline for robust malware detection.}\label{algo:systemAlgo}
\end{algorithm}

\subsection{Robust Malware Detection}\label{sec:handsonexamples}
In this section, we explore implementing a robust and accurate malware detection system using the following robustness guidelines in the development of features and the learning process.

\subsubsection{Robust Malware Representation: A Definition}\label{sec:robust_req}
For effective malware detection under binary-level section injection, the feature space should be under one of the following categories:

\begin{itemize}
    \item \textit{Independent Nonvolatile Features:} Features that (i) can not be modified arbitrarily, including the software certificate information and the sections start location, and (ii) can not be changed by modifying other arbitrary features, unlike op-codes histogram, can be leveraged for robust malware detection.
    
    \item \textit{Monotonically Increasing Accumulative Features:} In such features, adding content and information, caused by binary-level mutations, only results in increasing their vector representation values. These features can be utilized for malware detection, where the model learns a threshold, in which, if the software exceeds the threshold, the software is identified as malware. 
    
    \item \textit{Component-based Locally Computed Features:} Components-based features are features locally calculated from independent components, such as program sections. For robust malware detection, these locally computed features should correspond to a non-shared fixed location in the feature space. For instance, let $X$ be a software, $X'$ be the same software after binary-level mutation ($P$), and $f(.)$ is the feature representation of the software. Robust component-based feature representation holds under the following:
    $$
        X' = X \cup P \rightarrow  f(X') = f(X) \cup f(P).
    $$
    In this representation, binary-level mutations only perturb part of the feature vector, leaving the already existing patterns intact.
\end{itemize}

\begin{figure*}[t]
\centering
\includegraphics[width=0.8\textwidth]{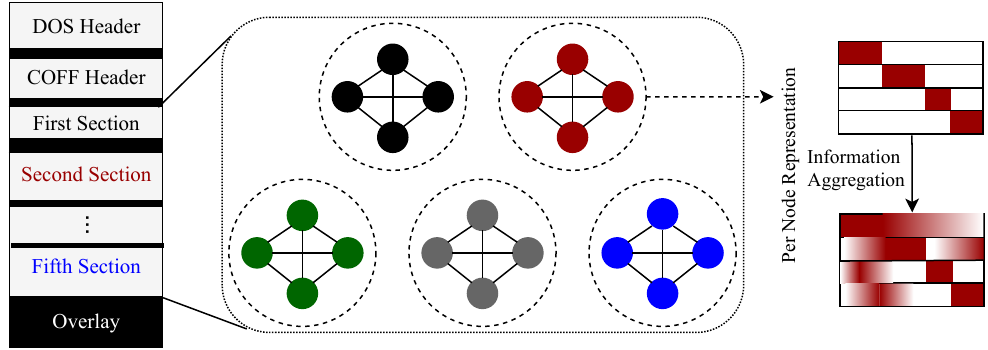}\vspace{-3mm}
\caption{Component-based graph representation. The different sections of the software are represented as isolated subgraphs in the feature space. Each subgraph consists of several nodes, each of which represents the section in different aspects (\ie bytes histogram, imports and function calls, or strings). Then, leveraging graph attention techniques, the information is aggregated between the nodes of the same section. }
\label{fig:InformationAgregation}\vspace{-5mm}
\end{figure*}

\subsubsection{Robust Learning Process: A Definition} For robust malware detection, the aforementioned feature categories should be accompanied by learning malicious informative patterns. Benign is diverse, and not bounded by existing patterns. On the other hand, malware categories (\ie families) are bounded by shared behavioral functionalities. The robust learning process is limited to only learning the malicious patterns, and not over-fit on the benign ``over-fitted'' patterns. The learning algorithm should only learn patterns associated with existing information, rather than patterns reflecting ``no information'' or ``absence of information''. This requires leveraging monotonic learning, and assigning a unique identifier for missing values (\ie ``NaN'') to prevent learning patterns reflecting \textit{absence of information}.

\subsubsection{System Design}\label{sec:sysDesignRobust}
The overall system design is shown in~\autoref{fig:SystemDesign}, and consists of four modules: software pre-processing, feature extraction, component-based graph encoding, and malware detection. The feature extraction and graph representation pipeline is shown in~\autoref{algo:systemAlgo}. In the following, we discuss each module.

\BfPara{Software Pre-processing}
In this step, we clean the software from any potential information injection and manipulation that can be done on binary-level, limiting the capabilities of the adversary. We follow the software pre-processing steps mentioned in~\autoref{sec:preprocessing}.

\BfPara{Feature Extraction}
After pre-processing, we extract robust features that fall into one of the three aforementioned categories (see~\autoref{sec:robust_req}). In particular, two types of features are used for the extraction process:

\noindent\textit{\underline{Component-based Features:}} Features related to independent components (i.e., software sections). Each component is represented using three different software representations, including:

\begin{itemize}
    \item \textit{Component's Bytes n-gram:} This representation transforms the bytes sequences into the natural language processing domain. Considering each byte (value from $0x00$ to $0xFF$) as a word in a long paragraph, we find the bytes n-grams with sizes 2--5 among the malware samples in the training dataset. Then, the extracted n-grams are represented as a sparse n-gram counts vector. 
    \item \textit{Component's Bytes Histogram:} The byte histogram contains 256 integer values, representing the counts of each byte value within the component. 
    
    \item \textit{Component's Strings:} The sequence of five or more characters in the range $0x20$ to $0x7f$ are considered as printable strings. The software is then represented as the count vectorizer of its printable strings.
\end{itemize}

\noindent\textit{\underline{Monotonically Increasing Features:}} Features that can not be arbitrarily modified, and the addition of information only leads to their values to increase. In the following, we describe the four different monotonically increasing representations utilized in this study:

\begin{itemize}
    \item \textit{Bytes Histogram:} Similar to the aforementioned component's histogram, however, it is calculated on the software level. 
    
    \item \textit{Software Imports:} The imports table is parsed to extract the imported functions and libraries. Then, the unique functions and libraries are represented using a count vectorizer.
    \item \textit{Software Exports:} Similar to the software imports, the exported functions are represented for the learning task.
    \item \textit{Software Strings:} Similar to the component's strings, but calculated on the software level.
\end{itemize}

\BfPara{Component-based Graph Encoding}
This module is only required for the component-based features. The monotonically increasing features are directly forwarded to the fourth module, while graph-based encoding techniques are utilized for the component's features.~\autoref{fig:InformationAgregation} shows the overview of this module process. 

\begin{enumerate}
    \item Each component (\ie section) is represented using different features associated with the functionality/behavior, including strings, and bytes n-gram.
    \item Each feature representation is then illustrated as a node in a graph space. Different nodes that represent the same component are fully connected, and there is no connection between nodes of different components. 
    \item Graph Attention Network is then used to encode the generated graph into a vector representation. The attention network is trained on malware detection task with non-negativity constraints.
    \item Once trained, we extract the graph embeddings vector representation after information aggregation (\ie after the attention operation). This allows for different representations to exchange essential information within the same component, increasing the malware detection accuracy. 
\end{enumerate}

\begin{table*}[h]
\centering
\caption{The baseline comparison results for malware detection task at 1\% and 3\% false positive rates (\ie misclassified benign software). PR: Padding Removal, SS: Software Stripping, BR: Bytes Resetting, Mod.: information and bytes values manipulation.}
\label{tab:baselinecomparison}\vspace{-3mm}
\scalebox{0.82}{
\begin{tabular}{!{\vrule width 1.2pt}l!{\vrule width 1.2pt}c|c|c!{\vrule width 1.2pt}c|c|c!{\vrule width 1.2pt}c|c|c!{\vrule width 1.2pt}c|c|c!{\vrule width 1.2pt}c|c|c!{\vrule width 1.2pt}c|c|c!{\vrule width 1.2pt}}
\Xhline{3\arrayrulewidth}
    \multirow{3}{*}{Baseline}  & \multicolumn{3}{c!{\vrule width 1.2pt}}{\multirow{2}{*}{Pre-processing}} & \multicolumn{15}{c!{\vrule width 1.2pt}}{Performance (TPR \% / FPR)}\\\cline{5-19} 
    & \multicolumn{3}{c!{\vrule width 1.2pt}}{} & \multicolumn{3}{c!{\vrule width 1.2pt}}{Baseline} & \multicolumn{3}{c!{\vrule width 1.2pt}}{Header Mod.} & \multicolumn{3}{c!{\vrule width 1.2pt}}{Binary Padding} & \multicolumn{3}{c!{\vrule width 1.2pt}}{Inter-section Mod.} & \multicolumn{3}{c!{\vrule width 1.2pt}}{Section Injection}\\\cline{2-19} 
& PR  & SS & BR & AUC & 1\% & 3\% & AUC &  1\% & 3\% & AUC &  1\% & 3\% & AUC &  1\% & 3\%  & AUC &  1\% & 3\% \\
\Xhline{2\arrayrulewidth}
\multirow{5}{*}{Ember} & \ecir &  \ecir &  \ecir & 91.15 & 54.46 & 61.56 & 79.81 & 7.30 & 18.30 &  74.87  &  2.37  &  5.37     & 91.17  &  52.92  &  61.72     &  77.21  &  30.73  & 37.87  \\\cline{2-19} 
 & \fcir &  \ecir &  \ecir & 90.21 & 50.63 & 58.97 & 77.59 & 6.50 & 11.66 & 90.21 & 50.63 & 58.97 &  88.83  &  46.78  &  53.08  &  79.75 & 39.00  & 48.11\\\cline{2-19} 
 & \ecir &  \fcir &  \ecir & 88.53 & 33.20 & 46.83 &  88.53 & 33.20 & 46.83 &   62.89  &  0.20  &  0.50 &  87.02  &  35.65  &  42.75  &  77.17 & 17.78  & 32.81\\\cline{2-19} 
 & \ecir &  \ecir &  \fcir & 90.77 & 52.98 & 60.48 & 78.54 & 5.63 & 8.76 &  73.20  &  1.00  &  4.90 &  90.77  &  50.88  &  60.31  & 75.90  &  27.25 & 34.98\\\cline{2-19} 
 & \fcir &  \fcir &  \fcir & 87.82 & 24.53 & 36.28 & 87.82 & 24.53 & 36.28  & 87.82 & 24.53 & 36.28   &  85.04  &  5.70  &  34.78  &  77.22 & 17.46 & 28.93\\\Xhline{3\arrayrulewidth}
\multirow{5}{*}{Ember (Mono)} & \ecir &  \ecir &  \ecir & 87.07  &  33.70  &  44.30 &  83.87  &  29.03  &  42.93  &   87.07  &  33.70  &  44.30  &    86.35  &  33.68  &  44.18 &   73.60 &  17.77  &  23.70 \\\cline{2-19} 
 & \fcir &  \ecir &  \ecir &  84.98  &  28.50  &  39.75  &  84.94  &  29.30  &  39.77  & 84.98  &  28.50  &  39.75  &  84.54  &  18.97  &  41.95 &   83.37  &  27.12  &  36.75  \\\cline{2-19} 
 & \ecir &  \fcir &  \ecir &  85.33  &  31.77  &  39.33 & 85.33  &  31.77  &  39.33   &   85.33  &  31.77  &  39.33     &  85.16  &  23.51  &  42.45  &  83.07  &  29.31  &  36.88 \\\cline{2-19} 
 & \ecir &  \ecir &  \fcir & 87.43  &  33.54  &  42.51  &   83.41  &  32.47  &  38.53  & 87.43  &  33.54  &  42.51   &  87.43  &  33.54  &  42.51  &    75.23  &  19.04  &  24.11 \\\cline{2-19} 
 & \fcir &  \fcir &  \fcir &  85.44  &  27.87  &  39.45 &  85.44  &  27.87  &  39.45   &  85.44  &  27.87  &  39.45   &   85.44  &  27.87  &  39.45   &   85.04  &  26.74  &  38.0 \\\Xhline{3\arrayrulewidth}
 \multirow{5}{*}{Software Visualization} & \ecir &  \ecir &  \ecir & 89.8  &  52.17  &  58.09  &  87.02  &  46.53  &  53.43 &  44.59  &  0.8  &  1.80  &  90.13  &  52.32  &  58.39  &  79.97  &  31.63  &  35.87\\\cline{2-19} 
 & \fcir &  \ecir &  \ecir & 89.01  &  49.4  &  53.91 & 88.06  &  49.53  &  54.06  &   89.01  &  49.4  &  53.91 &  88.16  &  37.78  &  52.72   & 88.03  &  44.83  &  49.60 \\\cline{2-19} 
 & \ecir &  \fcir &  \ecir &  88.41  &  41.57  &  51.83 & 88.41  &  41.57  &  51.83 & 33.23  &  0.00  &  0.13  &  87.67  &  35.71  &  50.78   &  86.35  &  38.27  &  47.70\\\cline{2-19} 
 & \ecir &  \ecir &  \fcir & 90.47  &  53.02  &  60.05 & 86.26  &  46.67  &  53.56  & 46.69  &  0.90  &  2.47 &   90.47  &  53.02  &  60.05  &  80.13  &  33.81  &  38.91
\\\cline{2-19} 
 & \fcir &  \fcir &  \fcir & 89.29  &  45.46  &  54.41 &  89.29  &  45.46  &  54.41  & 89.29  &  45.46  &  54.41  & 89.29  &  45.46  &  54.41   &  88.59  &  42.12  &  52.80 \\\Xhline{3\arrayrulewidth}
\multirow{5}{*}{MalConv} & \ecir &  \ecir &  \ecir &  88.47  &  46.93  &  56.67 &  84.02  &  27.37  &  34.37 & 89.42  &  47.90  &  58.57 & 88.21  &  46.62  &  55.95  &  82.53 & 31.19 & 51.70 \\\cline{2-19} 
 & \fcir &  \ecir &  \ecir &  88.2  &  49.37  &  57.04  & 82.57  &  21.33  &  36.17 & 88.2  &  49.37  &  57.04  &  88.03  &  44.51  &  55.98   &   82.43 & 34.62 & 44.14\\\cline{2-19} 
 & \ecir &  \fcir &  \ecir & 87.01  &  33.97  &  45.80 &  87.01  &  33.97  &  45.80 & 87.82  &  32.73  &  46.27  &  73.15  &  17.51  &  25.14   &  81.13 & 22.03 & 33.13 \\\cline{2-19} 
 & \ecir &  \ecir &  \fcir & 88.45  &  47.05  &  54.88 &  84.48  &  22.5  &  33.23 &  89.24  &  47.72  &  56.12  &  88.45  &  47.05  &  54.88  &  82.80 & 31.91  &  39.68\\\cline{2-19} 
 & \fcir &  \fcir &  \fcir &  87.04  &  40.75  &  50.90 &  87.04  &  40.75  &  50.90   &  87.04  &  40.75  &  50.90   &   87.04  &  40.75  &  50.90   & 81.36 & 31.51  & 44.26\\\Xhline{3\arrayrulewidth}
 \multirow{5}{*}{Online Engine --- 1} & \ecir &  \ecir &  \ecir &  \multicolumn{3}{c!{\vrule width 1.2pt}}{21.38} &  \multicolumn{3}{c!{\vrule width 1.2pt}}{25.07} &  \multicolumn{3}{c!{\vrule width 1.2pt}}{17.33}   &  \multicolumn{3}{c!{\vrule width 1.2pt}}{17.70}   &  \multicolumn{3}{c!{\vrule width 1.2pt}}{15.06} \\\cline{2-19} 
 & \fcir &  \ecir &  \ecir & \multicolumn{3}{c!{\vrule width 1.2pt}}{30.99} &  \multicolumn{3}{c!{\vrule width 1.2pt}}{27.30}  &  \multicolumn{3}{c!{\vrule width 1.2pt}}{30.99}  &  \multicolumn{3}{c!{\vrule width 1.2pt}}{31.29}   &  \multicolumn{3}{c!{\vrule width 1.2pt}}{15.62}\\\cline{2-19} 
 & \ecir &  \fcir &  \ecir &   \multicolumn{3}{c!{\vrule width 1.2pt}}{25.07} &  \multicolumn{3}{c!{\vrule width 1.2pt}}{25.07}  &  \multicolumn{3}{c!{\vrule width 1.2pt}}{18.60}  &  \multicolumn{3}{c!{\vrule width 1.2pt}}{22.79}   &  \multicolumn{3}{c!{\vrule width 1.2pt}}{14.96}\\\cline{2-19} 
 & \ecir &  \ecir &  \fcir &  \multicolumn{3}{c!{\vrule width 1.2pt}}{21.72} &  \multicolumn{3}{c!{\vrule width 1.2pt}}{31.04}  &  \multicolumn{3}{c!{\vrule width 1.2pt}}{17.05}  &  \multicolumn{3}{c!{\vrule width 1.2pt}}{21.72}   &  \multicolumn{3}{c!{\vrule width 1.2pt}}{14.77}\\\cline{2-19} 
 & \fcir &  \fcir &  \fcir & \multicolumn{3}{c!{\vrule width 1.2pt}}{31.58} &  \multicolumn{3}{c!{\vrule width 1.2pt}}{31.58}  &  \multicolumn{3}{c!{\vrule width 1.2pt}}{31.58}  &  \multicolumn{3}{c!{\vrule width 1.2pt}}{31.58}   &  \multicolumn{3}{c!{\vrule width 1.2pt}}{14.45}\\\Xhline{3\arrayrulewidth}
 \multirow{5}{*}{Online Engine --- 2} & \ecir &  \ecir &  \ecir &  \multicolumn{3}{c!{\vrule width 1.2pt}}{89.32} & \multicolumn{3}{c!{\vrule width 1.2pt}}{87.56} & \multicolumn{3}{c!{\vrule width 1.2pt}}{75.66}  & \multicolumn{3}{c!{\vrule width 1.2pt}}{82.15} & \multicolumn{3}{c!{\vrule width 1.2pt}}{69.01}\\\cline{2-19} 
 & \fcir &  \ecir &  \ecir & \multicolumn{3}{c!{\vrule width 1.2pt}}{84.62} &  \multicolumn{3}{c!{\vrule width 1.2pt}}{84.19}  &  \multicolumn{3}{c!{\vrule width 1.2pt}}{84.62}  &  \multicolumn{3}{c!{\vrule width 1.2pt}}{82.82}   &  \multicolumn{3}{c!{\vrule width 1.2pt}}{69.82}\\\cline{2-19}
 & \ecir &  \fcir &  \ecir &   \multicolumn{3}{c!{\vrule width 1.2pt}}{87.56} &  \multicolumn{3}{c!{\vrule width 1.2pt}}{87.56}  &  \multicolumn{3}{c!{\vrule width 1.2pt}}{61.20}  &  \multicolumn{3}{c!{\vrule width 1.2pt}}{82.00}   &  \multicolumn{3}{c!{\vrule width 1.2pt}}{76.46}\\\cline{2-19}
 & \ecir &  \ecir &  \fcir &  \multicolumn{3}{c!{\vrule width 1.2pt}}{88.54} &  \multicolumn{3}{c!{\vrule width 1.2pt}}{87.02}  &  \multicolumn{3}{c!{\vrule width 1.2pt}}{75.90}  &  \multicolumn{3}{c!{\vrule width 1.2pt}}{88.54}   &  \multicolumn{3}{c!{\vrule width 1.2pt}}{69.32}\\\cline{2-19} 
 & \fcir &  \fcir &  \fcir & \multicolumn{3}{c!{\vrule width 1.2pt}}{83.68} &  \multicolumn{3}{c!{\vrule width 1.2pt}}{83.68}  &  \multicolumn{3}{c!{\vrule width 1.2pt}}{83.68}  &  \multicolumn{3}{c!{\vrule width 1.2pt}}{83.68}   &  \multicolumn{3}{c!{\vrule width 1.2pt}}{76.70}\\\Xhline{3\arrayrulewidth}
 \multirow{5}{*}{Online Engine --- 3} & \ecir &  \ecir &  \ecir &  \multicolumn{3}{c!{\vrule width 1.2pt}}{73.19} &  \multicolumn{3}{c!{\vrule width 1.2pt}}{83.41} &  \multicolumn{3}{c!{\vrule width 1.2pt}}{63.30}  &  \multicolumn{3}{c!{\vrule width 1.2pt}}{69.38} &  \multicolumn{3}{c!{\vrule width 1.2pt}}{14.26}\\\cline{2-19} 
 & \fcir &  \ecir &  \ecir & \multicolumn{3}{c!{\vrule width 1.2pt}}{76.16} &  \multicolumn{3}{c!{\vrule width 1.2pt}}{76.00}  &  \multicolumn{3}{c!{\vrule width 1.2pt}}{76.16}  &  \multicolumn{3}{c!{\vrule width 1.2pt}}{76.07}   &  \multicolumn{3}{c!{\vrule width 1.2pt}}{10.61}\\\cline{2-19} 
 & \ecir &  \fcir &  \ecir &   \multicolumn{3}{c!{\vrule width 1.2pt}}{83.41} &  \multicolumn{3}{c!{\vrule width 1.2pt}}{83.41}  &  \multicolumn{3}{c!{\vrule width 1.2pt}}{57.03}  &  \multicolumn{3}{c!{\vrule width 1.2pt}}{68.56}   &  \multicolumn{3}{c!{\vrule width 1.2pt}}{10.60}\\\cline{2-19}  
 & \ecir &  \ecir &  \fcir &  \multicolumn{3}{c!{\vrule width 1.2pt}}{72.87} &  \multicolumn{3}{c!{\vrule width 1.2pt}}{77.59}  &  \multicolumn{3}{c!{\vrule width 1.2pt}}{62.63}  &  \multicolumn{3}{c!{\vrule width 1.2pt}}{72.87}   &  \multicolumn{3}{c!{\vrule width 1.2pt}}{13.50}\\\cline{2-19} 
 & \fcir &  \fcir &  \fcir & \multicolumn{3}{c!{\vrule width 1.2pt}}{75.21} &  \multicolumn{3}{c!{\vrule width 1.2pt}}{75.21}  &  \multicolumn{3}{c!{\vrule width 1.2pt}}{75.21}  &  \multicolumn{3}{c!{\vrule width 1.2pt}}{75.21}   &  \multicolumn{3}{c!{\vrule width 1.2pt}}{10.94}\\\Xhline{3\arrayrulewidth}
 \multirow{5}{*}{Online Engine --- 4} & \ecir &  \ecir &  \ecir &  \multicolumn{3}{c!{\vrule width 1.2pt}}{73.15} &   \multicolumn{3}{c!{\vrule width 1.2pt}}{83.01}  &   \multicolumn{3}{c!{\vrule width 1.2pt}}{59.56}  &   \multicolumn{3}{c!{\vrule width 1.2pt}}{69.01}   &   \multicolumn{3}{c!{\vrule width 1.2pt}}{48.06}\\\cline{2-19} 
 & \fcir &  \ecir &  \ecir & \multicolumn{3}{c!{\vrule width 1.2pt}}{79.95} &  \multicolumn{3}{c!{\vrule width 1.2pt}}{80.99}  &  \multicolumn{3}{c!{\vrule width 1.2pt}}{79.95}  &  \multicolumn{3}{c!{\vrule width 1.2pt}}{74.67}   &  \multicolumn{3}{c!{\vrule width 1.2pt}}{48.03}\\\cline{2-19} 
 & \ecir &  \fcir &  \ecir &   \multicolumn{3}{c!{\vrule width 1.2pt}}{83.01} &  \multicolumn{3}{c!{\vrule width 1.2pt}}{83.01}  &  \multicolumn{3}{c!{\vrule width 1.2pt}}{57.36}  &  \multicolumn{3}{c!{\vrule width 1.2pt}}{79.08}   &  \multicolumn{3}{c!{\vrule width 1.2pt}}{48.33}\\\cline{2-19}  
 & \ecir &  \ecir &  \fcir &  \multicolumn{3}{c!{\vrule width 1.2pt}}{72.42} &  \multicolumn{3}{c!{\vrule width 1.2pt}}{75.63}  &  \multicolumn{3}{c!{\vrule width 1.2pt}}{59.22}  &  \multicolumn{3}{c!{\vrule width 1.2pt}}{72.42}   &  \multicolumn{3}{c!{\vrule width 1.2pt}}{47.98}\\\cline{2-19} 
 & \fcir &  \fcir &  \fcir & \multicolumn{3}{c!{\vrule width 1.2pt}}{78.92} &  \multicolumn{3}{c!{\vrule width 1.2pt}}{78.92}  &  \multicolumn{3}{c!{\vrule width 1.2pt}}{78.92}  &  \multicolumn{3}{c!{\vrule width 1.2pt}}{78.92}   &  \multicolumn{3}{c!{\vrule width 1.2pt}}{48.19}\\\Xhline{3\arrayrulewidth}
 \multirow{5}{*}{Online Engine --- 5} & \ecir &  \ecir &  \ecir &  \multicolumn{3}{c!{\vrule width 1.2pt}}{82.22} &   \multicolumn{3}{c!{\vrule width 1.2pt}}{87.06}  &   \multicolumn{3}{c!{\vrule width 1.2pt}}{69.20} &   \multicolumn{3}{c!{\vrule width 1.2pt}}{77.38}  &   \multicolumn{3}{c!{\vrule width 1.2pt}}{60.93}\\\cline{2-19} 
 & \fcir &  \ecir &  \ecir & \multicolumn{3}{c!{\vrule width 1.2pt}}{90.24} &  \multicolumn{3}{c!{\vrule width 1.2pt}}{91.72}  &  \multicolumn{3}{c!{\vrule width 1.2pt}}{90.24}  &  \multicolumn{3}{c!{\vrule width 1.2pt}}{91.25}   &  \multicolumn{3}{c!{\vrule width 1.2pt}}{56.50}\\\cline{2-19} 
 & \ecir &  \fcir &  \ecir &   \multicolumn{3}{c!{\vrule width 1.2pt}}{87.06} &  \multicolumn{3}{c!{\vrule width 1.2pt}}{87.06}  &  \multicolumn{3}{c!{\vrule width 1.2pt}}{49.83}  &  \multicolumn{3}{c!{\vrule width 1.2pt}}{82.10}   &  \multicolumn{3}{c!{\vrule width 1.2pt}}{56.93}\\\cline{2-19} 
 & \ecir &  \ecir &  \fcir &  \multicolumn{3}{c!{\vrule width 1.2pt}}{83.51} &  \multicolumn{3}{c!{\vrule width 1.2pt}}{86.11}  &  \multicolumn{3}{c!{\vrule width 1.2pt}}{69.00}  &  \multicolumn{3}{c!{\vrule width 1.2pt}}{83.51}   &  \multicolumn{3}{c!{\vrule width 1.2pt}}{62.02}\\\cline{2-19} 
 & \fcir &  \fcir &  \fcir & \multicolumn{3}{c!{\vrule width 1.2pt}}{92.61} &  \multicolumn{3}{c!{\vrule width 1.2pt}}{92.61}  &  \multicolumn{3}{c!{\vrule width 1.2pt}}{92.61}  &  \multicolumn{3}{c!{\vrule width 1.2pt}}{92.61}   &  \multicolumn{3}{c!{\vrule width 1.2pt}}{58.91}\\\Xhline{3\arrayrulewidth}
 
\multirow{5}{*}{Graph Encoding (Ours)}  & \ecir &  \ecir &  \ecir & 88.32  &  50.50  &  56.08 &  86.96  &  48.59  &  53.32 & 88.32  &  50.50  &  56.08  &  87.91  &  50.20  &  55.88   &  87.06 & 48.93  & 54.27\\\cline{2-19} 
 & \fcir &  \ecir &  \ecir & 88.44  &  51.08  &  54.51 & 86.88  &  47.19  &  51.89  & 88.44  &  51.08  &  54.51 & 88.06  &  51.26  &  54.22   & 87.73  &  50.60 & 53.64\\\cline{2-19} 
 & \ecir &  \fcir &  \ecir & 88.45  &  50.02  &  56.58 &  88.45  &  50.02  &  56.58  & 88.45  &  50.02  &  56.58 &   87.28  &  49.22  &  52.79   &  87.91 & 48.78  & 51.36\\\cline{2-19} 
 & \ecir &  \ecir &  \fcir & 88.48  &  48.37  &  54.97 &  87.09  &  46.24  &  50.4  &  88.48  &  48.37  &  54.97  &   88.48  &  48.37  &  54.97   &   88.07 & 48.21  &  54.23\\\cline{2-19} 
 & \fcir &  \fcir &  \fcir &  88.89  &  51.74  &  55.59 &  88.89  &  51.74  &  55.59 & 88.89  &  51.74  &  55.59   &   88.89  &  51.74  &  55.59   &   88.19  &  51.33  &  54.81 \\

\Xhline{3\arrayrulewidth}

\end{tabular}}\vspace{-2mm}

\end{table*}

\noindent\textit{\underline{Why Graph Attention?}}
Traditionally, when considering different feature representations, two approaches can be utilized in the learning process: (i) concatenate the vectors of different feature representations toward malware detection task, or (ii) build an ensemble of malware detection classifiers, each of which is trained on specific feature representation. 

However, we argue that both approaches cause loss of information. For instance, concatenating the feature representations may result in fake dependencies between different representations' features among different components. Moreover, when using ensemble learning, each individual detection model is not aware of the other representations of the component, resulting in the loss of potential information and discriminative patterns. 

In this work, we address this issue by leveraging graph attention. Each node, associated with feature representation, will be updated by the attention process, with important information from other representations of the same component. Then, the per-node representation after information aggregation, referred to as encoding, is extracted and forwarded to the next module for malware detection. We note that the encoding of the graph is the combination of the per-node representations.

\BfPara{Malware Detection}
The monotonically increasing features and the extracted graph encoding are then forwarded to LightGBM~\cite{ke2017lightgbm} tree learning structure for the malware detection task.

\subsection{Evaluation Results}
We evaluated the robustness of five different malware detection models, including our proposed graph encoding-based malware detection scheme, and five online detection engines accessible through VirusTotal API. 
The evaluation results are shown in~\autoref{tab:baselinecomparison}.
Notice that our design out-perform the other baselines under false-positive rates of 1\% and 3\% on both clean and adversarial settings. For a better understanding of the robustness of the models and engines, each attack is discussed individually. 

\subsubsection{Baseline Performance Comparison} 
For each representation, we trained five baseline models based on (i) original binaries, and binaries after (ii) padding removal, (iii) software stripping, (iv) unmapped binaries resetting, and (v) combined software pre-processing. 
The EMBER-based model provides state-of-the-art detection performance, with an AUC score of 91.15\%. However, this performance deteriorates under robust software pre-processing, down to 87.82\%. This trend is more noticeable considering a benign false positive rate threshold of 1\%. On the other hand, our proposed model performance persists under the various software pre-processing techniques. 

Moreover, we notice that the five online engines' performance increase after pre-processing, with an exception of Online Engine -- 2. This is mainly attributed to the removal of adversarial channels that may be utilized for misclassification.

\noindent\textbf{\underline{\textit{Key Takeaway:}}} Removing volatile information channels through software pre-processing slightly decreases the performance of the implemented malware detection models. This is attributed to the removal of information that may be used for malicious patterns recognition. However, for large-scale online engines, the removal of these channels provides better detection accuracy. 

\subsubsection{Header Information Stripping} 
In these settings, we evaluate the aforementioned baseline models against software binaries under header modification (\ie stripping) attack. After stripping the header information of the test dataset, the software is pre-processed and evaluated against the baseline models. We note that benign samples were not modified. As shown in table~\ref{tab:baselinecomparison}, binary stripping reduces the accuracy of all implemented baseline models. This is more prevalent in the EMBER model, as the header information is directly used for malware detection. However, for monotonically increasing ember and graph encoding models, stripping the binaries affect the bytes distribution and histogram, resulting in a slight decrease in performance. However, under software stripping pre-processing, the attack channel is eliminated, and the models report exact performance on both baseline and header information modified binaries. 

On the online engines level, we notice that the detection performance of the engines increased, with the exception of the second engine, \textit{Online Engine -- 2}.  

\noindent\textbf{\underline{\textit{Key Takeaway:}}} Software stripping pre-processing eliminate the header manipulation attack channel, invalidating its effects on the baseline detection accuracy.

\subsubsection{Benign Binaries Padding} 
Padding benign binaries at the end of the test dataset malicious binaries highly decreases the detection performance of both EMBER and software visualization-based malware detection models. However, for both the monotonically increasing EMBER model and graph encoding model, the performance before and after binary padding is similar, as addition attacks do not affect the performance of monotonically increasing models. 

Additionally, binary padding highly reduces the performance of online detection engines, with up to 38\% loss of detection performance.
However, under padding removal, the attack channel is invalidated, and the performance of the models is similar to their baseline counterparts.  

\noindent\textbf{\underline{\textit{Key Takeaway:}}} Binary padding significantly reduce the detection performance of various models and engines. However, using padding removal pre-processing eliminates its adversarial effects.

\subsubsection{Inter-section Injection} In this setting, we randomize the unmapped bytes between different sections of the software. The surface of this attack is limited by design, and does not cause a high reduction in performance, as shown in table~\ref{tab:baselinecomparison}. 
However, it is essential to understand and eliminate this attack channel, as malware authors may opt for artificially increasing the space between sections to increase the attack surface. 
This can be addressed by following inter-section bytes resetting, which eliminates this attack channel, mitigating its effects.

\noindent\textbf{\underline{\textit{Key Takeaway:}}} Inter-section bytes resetting effectively eliminates the effects of inter-section injection, including potential code caves and targeted adversarial binaries.

\subsubsection{Benign Section Injection} 
Section injection is an emerging attack, where new sections are added to the binaries, with no limitations on the number of sections, nor their sizes. In this setting, benign sections are injected into each malicious software in the test dataset, until having a total of ten sections in the software. Notice that the performance of all the models was highly reduced, with the exception of the graph encoding baseline model. This is mainly resulted due to that the graph encoding model considers a section-based representation, while other representations consider the software as a whole. We recall that adding a section will also result in overwriting existing unmapped bytes, which reduces the performance of the monotonically increasing EMBER model. 

The same trend can be observed by the online engines, where all engines reduce detection performance after injecting the benign sections. It is worth mentioning that, despite the fact that software pre-processing mitigates all previous adversarial attacks settings, the newly added sections are part of the executable, and can not be directly identified using existing pre-processing and static analysis techniques. 

\noindent\textbf{\underline{\textit{Key Takeaway:}}} Section injection is effective against both online detection engines and the implemented malware detection models. This attack, in contrast to the previous attacks, can not be mitigated using software pre-processing. However, using section-based representation, accompanied by monotonic learning, we were able to effectively detect the adversarial malware samples with a reduced detection performance trade-off. 

We refer the reader to section~\ref{sec:classification_task_analysis} in the appendix for further information regarding the classification task impact of utilizing Graph Encoding on the reported accuracy.

\section{Discussion \& Limitations}\label{sec:discussion}

\BfPara{Binary-level Malware Detection}
Malware detection using binary-level features is challenging. A malware binary representation can significantly vary based on the compiler and underneath architecture. Despite that, with the rapid growth of malware mutations, it is considered the de facto approach for end-point devices malware detection. In this work, we tackle the challenge of binary-level malware mutations, providing pre-processing techniques, alongside a novel graph encoding approach for robust malware detection. 
In particular, the proposed approach is robust against header information removal, binary passing, inter-section gap modification, and section injection attacks. All these attacks are leveraged on the binary-level to generate malware mutations with different signatures to bypass traditional malware detection approaches. 

\BfPara{Binary Packing}
Binary packing is a common technique for malware obfuscation, in which, malware binaries are compressed or transformed. Upon starting the execution, the packer utility starts with de-obfuscating the malware, then executing it. Binary packing has been shown to bypass all binary-level malware detectors~\cite{abusnaina2022systematically}, where all obfuscated samples are classified as benign, or malware, regardless of their behavior. Due to the dependency on the binary packing vulnerable features, the proposed graph encoding technique also suffers from the same caveats. 

\BfPara{Black-box Adversarial Attacks} In this work, we focus on adversarial attacks that are (i) unaware of the underneath malware detection system, (ii) can be conducted on the end-point device without third-party software. We acknowledge the existence of various black-box malware detection attacks beyond the ones discussed in this work. However, we argue that, without having access to the code base, any modifications that assume adversarial example excitability are limited to the same channels targeted in section~\ref{sec:adversarial_capabilities}. While this work doesn't cover a variety of adversarial attacks, the complete removal of information dependency within these channels can be generalized to any attack targeting the channels.

\section{Concluding Remarks}\label{sec:conclusion}
In this work, we revisit the adversarial capabilities under binary-level mutations, highlighting the potential exploitation of implementing lightweight malware defenses on end-point devices. Addressing this critical issue, we investigate the vulnerabilities within existing malware detection approaches, formulating guidelines for robust malware detection. Further, we proposed simple-yet-effective software pre-processing techniques for robust malware detection and adversarial mitigation. 

While these techniques are capable of eliminating several attack channels, they are still vulnerable to the emerging section injection attack. Toward this, we proposed the first machine learning-based robust malware detection framework that encodes program's sections into multi-graph attention encoding for malware detection. The extracted encoding, accompanied by the monotonically increasing features leveraged in the literature, are used to achieve state-of-the-art robustness-to-performance trade-off. We further show that the proposed pre-processing techniques and graph encoding are indeed capturing and preserving per malware family capabilities, enabling robust malware detection and classification.


\balance
\bibliographystyle{ACM-Reference-Format}
\bibliography{ref}


\begin{thebibliography}{48}


\ifx \showCODEN    \undefined \def \showCODEN     #1{\unskip}     \fi
\ifx \showDOI      \undefined \def \showDOI       #1{#1}\fi
\ifx \showISBNx    \undefined \def \showISBNx     #1{\unskip}     \fi
\ifx \showISBNxiii \undefined \def \showISBNxiii  #1{\unskip}     \fi
\ifx \showISSN     \undefined \def \showISSN      #1{\unskip}     \fi
\ifx \showLCCN     \undefined \def \showLCCN      #1{\unskip}     \fi
\ifx \shownote     \undefined \def \shownote      #1{#1}          \fi
\ifx \showarticletitle \undefined \def \showarticletitle #1{#1}   \fi
\ifx \showURL      \undefined \def \showURL       {\relax}        \fi
\providecommand\bibfield[2]{#2}
\providecommand\bibinfo[2]{#2}
\providecommand\natexlab[1]{#1}
\providecommand\showeprint[2][]{arXiv:#2}

\bibitem[Vir(2019)]%
        {VirusTotal}
 \bibinfo{year}{2019}\natexlab{}.
\newblock \bibinfo{title}{{VirusTotal}}.
\newblock \bibinfo{howpublished}{Available at [Online]:
  \url{https://www.virustotal.com}}.
\newblock


\bibitem[Vir(2021)]%
        {VirusTotalStats}
 \bibinfo{year}{2021}\natexlab{}.
\newblock \bibinfo{title}{{VirusTotal Statistics}}.
\newblock \bibinfo{howpublished}{Available at [Online]:
  \url{https://www.virustotal.com/en/statistics/}}.
\newblock


\bibitem[Vir(2022)]%
        {VirusShare}
 \bibinfo{year}{2022}\natexlab{}.
\newblock \bibinfo{title}{{V}irus{S}hare}.
\newblock \bibinfo{howpublished}{Available at [Online]:
  \url{https://virusshare.com/}}.
\newblock


\bibitem[Abusnaina et~al\mbox{.}(2019a)]%
        {AbusnainaAASNM19}
\bibfield{author}{\bibinfo{person}{Ahmed Abusnaina}, \bibinfo{person}{Hisham
  Alasmary}, \bibinfo{person}{Mohammed Abuhamad}, \bibinfo{person}{Saeed
  Salem}, \bibinfo{person}{DaeHun Nyang}, {and} \bibinfo{person}{Aziz
  Mohaisen}.} \bibinfo{year}{2019}\natexlab{a}.
\newblock \showarticletitle{Subgraph-Based Adversarial Examples Against
  Graph-Based IoT Malware Detection Systems}. In
  \bibinfo{booktitle}{\emph{International Conference on Computational Data and
  Social Networks}}. \bibinfo{pages}{268--281}.
\newblock


\bibitem[Abusnaina et~al\mbox{.}(2022)]%
        {abusnaina2022systematically}
\bibfield{author}{\bibinfo{person}{Ahmed Abusnaina}, \bibinfo{person}{Afsah
  Anwar}, \bibinfo{person}{Sultan Alshamrani}, \bibinfo{person}{Abdulrahman
  Alabduljabbar}, \bibinfo{person}{Rhongho Jang}, \bibinfo{person}{DaeHun
  Nyang}, {and} \bibinfo{person}{David Mohaisen}.}
  \bibinfo{year}{2022}\natexlab{}.
\newblock \showarticletitle{Systematically Evaluating the Robustness of
  ML-based IoT Malware Detection Systems}. In
  \bibinfo{booktitle}{\emph{Proceedings of the 25th International Symposium on
  Research in Attacks, Intrusions and Defenses}}. \bibinfo{pages}{308--320}.
\newblock


\bibitem[Abusnaina et~al\mbox{.}(2019b)]%
        {AbusnainaMYM19}
\bibfield{author}{\bibinfo{person}{Ahmed Abusnaina}, \bibinfo{person}{DaeHun
  Nyang}, \bibinfo{person}{Murat Yuksel}, {and} \bibinfo{person}{Aziz
  Mohaisen}.} \bibinfo{year}{2019}\natexlab{b}.
\newblock \showarticletitle{Examining the Security of DDoS Detection Systems in
  Software Defined Networks}. In \bibinfo{booktitle}{\emph{Proceedings of the
  15th International Conference on emerging Networking EXperiments and
  Technologies}}. \bibinfo{pages}{49--50}.
\newblock


\bibitem[Al{-}Dujaili et~al\mbox{.}(2018)]%
        {Al-DujailiHHO18}
\bibfield{author}{\bibinfo{person}{Abdullah Al{-}Dujaili},
  \bibinfo{person}{Alex Huang}, \bibinfo{person}{Erik Hemberg}, {and}
  \bibinfo{person}{Una{-}May O'Reilly}.} \bibinfo{year}{2018}\natexlab{}.
\newblock \showarticletitle{Adversarial Deep Learning for Robust Detection of
  Binary Encoded Malware}. In \bibinfo{booktitle}{\emph{Proceedings of the
  {IEEE} Security and Privacy Workshops, {SP} Workshops}}.
  \bibinfo{pages}{76--82}.
\newblock


\bibitem[Alasmary et~al\mbox{.}(2019a)]%
        {AlasmaryAPCNM19}
\bibfield{author}{\bibinfo{person}{Hisham Alasmary}, \bibinfo{person}{Aminollah
  Khormali}, \bibinfo{person}{Afsah Anwar}, \bibinfo{person}{Jeman Park},
  \bibinfo{person}{Jinchun Choi}, \bibinfo{person}{Ahmed Abusnaina},
  \bibinfo{person}{Amro Awad}, \bibinfo{person}{DaeHun Nyang}, {and}
  \bibinfo{person}{Aziz Mohaisen}.} \bibinfo{year}{2019}\natexlab{a}.
\newblock \showarticletitle{{Analyzing and Detecting Emerging Internet of
  Things Malware: A Graph-based Approach}}.
\newblock \bibinfo{journal}{\emph{IEEE Internet of Things Journal}}
  (\bibinfo{year}{2019}).
\newblock


\bibitem[Alasmary et~al\mbox{.}(2019b)]%
        {AlasmaryKAPCAANM19}
\bibfield{author}{\bibinfo{person}{Hisham Alasmary}, \bibinfo{person}{Aminollah
  Khormali}, \bibinfo{person}{Afsah Anwar}, \bibinfo{person}{Jeman Park},
  \bibinfo{person}{Jinchun Choi}, \bibinfo{person}{Ahmed Abusnaina},
  \bibinfo{person}{Amro Awad}, \bibinfo{person}{DaeHun Nyang}, {and}
  \bibinfo{person}{Aziz Mohaisen}.} \bibinfo{year}{2019}\natexlab{b}.
\newblock \showarticletitle{Analyzing and detecting emerging Internet of Things
  malware: a graph-based approach}.
\newblock \bibinfo{journal}{\emph{{IEEE} Internet of Things Journal}}
  \bibinfo{volume}{6}, \bibinfo{number}{5} (\bibinfo{year}{2019}),
  \bibinfo{pages}{8977--8988}.
\newblock


\bibitem[Alrawi et~al\mbox{.}(2021)]%
        {AlrawiLVSMAM21}
\bibfield{author}{\bibinfo{person}{Omar Alrawi}, \bibinfo{person}{Charles
  Lever}, \bibinfo{person}{Kevin Valakuzhy}, \bibinfo{person}{Kevin Snow},
  \bibinfo{person}{Fabian Monrose}, \bibinfo{person}{Manos Antonakakis},
  {et~al\mbox{.}}} \bibinfo{year}{2021}\natexlab{}.
\newblock \showarticletitle{The Circle Of Life: A Large-Scale Study of The IoT
  Malware Lifecycle}. In \bibinfo{booktitle}{\emph{{USENIX} Security Symposium
  ({USENIX} Security 21)}}.
\newblock


\bibitem[Anderson and Roth(2018)]%
        {anderson2018ember}
\bibfield{author}{\bibinfo{person}{Hyrum~S Anderson} {and}
  \bibinfo{person}{Phil Roth}.} \bibinfo{year}{2018}\natexlab{}.
\newblock \showarticletitle{Ember: an open dataset for training static pe
  malware machine learning models}.
\newblock \bibinfo{journal}{\emph{arXiv preprint arXiv:1804.04637}}
  (\bibinfo{year}{2018}).
\newblock


\bibitem[Anwar et~al\mbox{.}(2020)]%
        {AnwarAPW20}
\bibfield{author}{\bibinfo{person}{Afsah Anwar}, \bibinfo{person}{Hisham
  Alasmary}, \bibinfo{person}{Jeman Park}, \bibinfo{person}{An Wang},
  \bibinfo{person}{Songqing Chen}, {and} \bibinfo{person}{David Mohaisen}.}
  \bibinfo{year}{2020}\natexlab{}.
\newblock \showarticletitle{Statically Dissecting Internet of Things Malware:
  Analysis, Characterization, and Detection}. In
  \bibinfo{booktitle}{\emph{International Conference on Information and
  Communications Security}}. Springer, \bibinfo{pages}{443--461}.
\newblock


\bibitem[Barr-Smith et~al\mbox{.}(2021)]%
        {barr2021survivalism}
\bibfield{author}{\bibinfo{person}{Frederick Barr-Smith},
  \bibinfo{person}{Xabier Ugarte-Pedrero}, \bibinfo{person}{Mariano Graziano},
  \bibinfo{person}{Riccardo Spolaor}, {and} \bibinfo{person}{Ivan Martinovic}.}
  \bibinfo{year}{2021}\natexlab{}.
\newblock \showarticletitle{Survivalism: Systematic Analysis of Windows Malware
  Living-Off-The-Land}. In \bibinfo{booktitle}{\emph{Proceedings of the IEEE
  Symposium on Security and Privacy}}.
\newblock


\bibitem[Cai et~al\mbox{.}(2018)]%
        {CaiMRY18}
\bibfield{author}{\bibinfo{person}{Haipeng Cai}, \bibinfo{person}{Na Meng},
  \bibinfo{person}{Barbara Ryder}, {and} \bibinfo{person}{Daphne Yao}.}
  \bibinfo{year}{2018}\natexlab{}.
\newblock \showarticletitle{Droidcat: Effective android malware detection and
  categorization via app-level profiling}.
\newblock \bibinfo{journal}{\emph{IEEE Transactions on Information Forensics
  and Security}} \bibinfo{volume}{14}, \bibinfo{number}{6}
  (\bibinfo{year}{2018}), \bibinfo{pages}{1455--1470}.
\newblock


\bibitem[Carlini and Wagner(2017a)]%
        {Carlini017_2}
\bibfield{author}{\bibinfo{person}{Nicholas Carlini} {and}
  \bibinfo{person}{David~A. Wagner}.} \bibinfo{year}{2017}\natexlab{a}.
\newblock \showarticletitle{Adversarial Examples Are Not Easily Detected:
  Bypassing Ten Detection Methods}. In \bibinfo{booktitle}{\emph{Proceedings of
  the 10th {ACM} Workshop on Artificial Intelligence and Security, AISec@CCS}}.
  \bibinfo{pages}{3--14}.
\newblock


\bibitem[Carlini and Wagner(2017b)]%
        {Carlini017}
\bibfield{author}{\bibinfo{person}{Nicholas Carlini} {and}
  \bibinfo{person}{David~A. Wagner}.} \bibinfo{year}{2017}\natexlab{b}.
\newblock \showarticletitle{Towards Evaluating the Robustness of Neural
  Networks}. In \bibinfo{booktitle}{\emph{Proceedings of the {IEEE} Symposium
  on Security and Privacy}}. \bibinfo{pages}{39--57}.
\newblock


\bibitem[Castro et~al\mbox{.}(2019)]%
        {castro2019armed}
\bibfield{author}{\bibinfo{person}{Raphael~Labaca Castro},
  \bibinfo{person}{Corinna Schmitt}, {and} \bibinfo{person}{Gabi~Dreo
  Rodosek}.} \bibinfo{year}{2019}\natexlab{}.
\newblock \showarticletitle{Armed: How automatic malware modifications can
  evade static detection?}. In \bibinfo{booktitle}{\emph{2019 5th International
  Conference on Information Management (ICIM)}}. \bibinfo{pages}{20--27}.
\newblock


\bibitem[Cui et~al\mbox{.}(2018)]%
        {CuiXCCWC18}
\bibfield{author}{\bibinfo{person}{Zhihua Cui}, \bibinfo{person}{Fei Xue},
  \bibinfo{person}{Xingjuan Cai}, \bibinfo{person}{Yang Cao},
  \bibinfo{person}{Gaige Wang}, {and} \bibinfo{person}{Jinjun Chen}.}
  \bibinfo{year}{2018}\natexlab{}.
\newblock \showarticletitle{Detection of Malicious Code Variants Based on Deep
  Learning}.
\newblock \bibinfo{journal}{\emph{Trans. Industrial Informatics}}
  \bibinfo{volume}{14}, \bibinfo{number}{7} (\bibinfo{year}{2018}),
  \bibinfo{pages}{3187--3196}.
\newblock


\bibitem[Demetrio et~al\mbox{.}(2021)]%
        {DemetrioBLRA21}
\bibfield{author}{\bibinfo{person}{Luca Demetrio}, \bibinfo{person}{Battista
  Biggio}, \bibinfo{person}{Giovanni Lagorio}, \bibinfo{person}{Fabio Roli},
  {and} \bibinfo{person}{Alessandro Armando}.} \bibinfo{year}{2021}\natexlab{}.
\newblock \showarticletitle{Functionality-Preserving Black-Box Optimization of
  Adversarial Windows Malware}.
\newblock \bibinfo{journal}{\emph{{IEEE} Trans. Inf. Forensics Secur.}}
  \bibinfo{volume}{16} (\bibinfo{year}{2021}), \bibinfo{pages}{3469--3478}.
\newblock


\bibitem[Demetrio et~al\mbox{.}(2020)]%
        {DCBLAR2020}
\bibfield{author}{\bibinfo{person}{Luca Demetrio}, \bibinfo{person}{Scott~E.
  Coull}, \bibinfo{person}{Battista Biggio}, \bibinfo{person}{Giovanni
  Lagorio}, \bibinfo{person}{Alessandro Armando}, {and} \bibinfo{person}{Fabio
  Roli}.} \bibinfo{year}{2020}\natexlab{}.
\newblock \showarticletitle{Adversarial {EXE}mples: {A} Survey and Experimental
  Evaluation of Practical Attacks on Machine Learning for Windows Malware
  Detection}.
\newblock \bibinfo{journal}{\emph{CoRR}}  \bibinfo{volume}{abs/2008.07125}
  (\bibinfo{year}{2020}).
\newblock


\bibitem[Fu et~al\mbox{.}(2018)]%
        {FuXWLS18}
\bibfield{author}{\bibinfo{person}{Jianwen Fu}, \bibinfo{person}{Jingfeng Xue},
  \bibinfo{person}{Yong Wang}, \bibinfo{person}{Zhenyan Liu}, {and}
  \bibinfo{person}{Chun Shan}.} \bibinfo{year}{2018}\natexlab{}.
\newblock \showarticletitle{Malware Visualization for Fine-Grained
  Classification}.
\newblock \bibinfo{journal}{\emph{{IEEE} Access}}  \bibinfo{volume}{6}
  (\bibinfo{year}{2018}), \bibinfo{pages}{14510--14523}.
\newblock


\bibitem[Grosse et~al\mbox{.}(2017)]%
        {GrossePMBM17}
\bibfield{author}{\bibinfo{person}{Kathrin Grosse}, \bibinfo{person}{Nicolas
  Papernot}, \bibinfo{person}{Praveen Manoharan}, \bibinfo{person}{Michael
  Backes}, {and} \bibinfo{person}{Patrick~D. McDaniel}.}
  \bibinfo{year}{2017}\natexlab{}.
\newblock \showarticletitle{Adversarial Examples for Malware Detection}. In
  \bibinfo{booktitle}{\emph{Proceedings of the 22nd European Symposium on
  Research Computer Security - {ESORICS}, Part {II}}}. \bibinfo{pages}{62--79}.
\newblock


\bibitem[HaddadPajouh et~al\mbox{.}(2018)]%
        {haddadpajouh2018deep}
\bibfield{author}{\bibinfo{person}{Hamed HaddadPajouh}, \bibinfo{person}{Ali
  Dehghantanha}, \bibinfo{person}{Raouf Khayami}, {and}
  \bibinfo{person}{Kim-Kwang~Raymond Choo}.} \bibinfo{year}{2018}\natexlab{}.
\newblock \showarticletitle{A deep Recurrent Neural Network based approach for
  Internet of Things malware threat hunting}.
\newblock \bibinfo{journal}{\emph{Future Generation Computer Systems}}
  \bibinfo{volume}{85} (\bibinfo{year}{2018}), \bibinfo{pages}{88--96}.
\newblock


\bibitem[Incer et~al\mbox{.}(2018)]%
        {IncerTA018}
\bibfield{author}{\bibinfo{person}{Inigo Incer}, \bibinfo{person}{Michael
  Theodorides}, \bibinfo{person}{Sadia Afroz}, {and} \bibinfo{person}{David~A.
  Wagner}.} \bibinfo{year}{2018}\natexlab{}.
\newblock \showarticletitle{Adversarially Robust Malware Detection Using
  Monotonic Classification}. In \bibinfo{booktitle}{\emph{Proceedings of the
  Fourth {ACM} International Workshop on Security and Privacy Analytics,
  IWSPA@CODASPY 2018, Tempe, AZ, USA, March 19-21, 2018}},
  \bibfield{editor}{\bibinfo{person}{Rakesh~M. Verma} {and}
  \bibinfo{person}{Murat Kantarcioglu}} (Eds.). \bibinfo{publisher}{{ACM}},
  \bibinfo{pages}{54--63}.
\newblock


\bibitem[Jordaney et~al\mbox{.}(2017)]%
        {jordaney2017transcend}
\bibfield{author}{\bibinfo{person}{Roberto Jordaney}, \bibinfo{person}{Kumar
  Sharad}, \bibinfo{person}{Santanu~K Dash}, \bibinfo{person}{Zhi Wang},
  \bibinfo{person}{Davide Papini}, \bibinfo{person}{Ilia Nouretdinov}, {and}
  \bibinfo{person}{Lorenzo Cavallaro}.} \bibinfo{year}{2017}\natexlab{}.
\newblock \showarticletitle{Transcend: Detecting concept drift in malware
  classification models}. In \bibinfo{booktitle}{\emph{26th {USENIX} Security
  Symposium ({USENIX} Security 17)}}. \bibinfo{pages}{625--642}.
\newblock


\bibitem[Kang et~al\mbox{.}(2015)]%
        {kang2015detecting}
\bibfield{author}{\bibinfo{person}{Hyunjae Kang}, \bibinfo{person}{Jae-wook
  Jang}, \bibinfo{person}{Aziz Mohaisen}, {and} \bibinfo{person}{Huy~Kang
  Kim}.} \bibinfo{year}{2015}\natexlab{}.
\newblock \showarticletitle{Detecting and classifying android malware using
  static analysis along with creator information}.
\newblock \bibinfo{journal}{\emph{International Journal of Distributed Sensor
  Networks}} \bibinfo{volume}{11}, \bibinfo{number}{6} (\bibinfo{year}{2015}),
  \bibinfo{pages}{479174}.
\newblock


\bibitem[Kantchelian et~al\mbox{.}(2013)]%
        {KantchelianAHIMTGJT2013}
\bibfield{author}{\bibinfo{person}{Alex Kantchelian}, \bibinfo{person}{Sadia
  Afroz}, \bibinfo{person}{Ling Huang}, \bibinfo{person}{Aylin~Caliskan Islam},
  \bibinfo{person}{Brad Miller}, \bibinfo{person}{Michael~Carl Tschantz},
  \bibinfo{person}{Rachel Greenstadt}, \bibinfo{person}{Anthony~D Joseph},
  {and} \bibinfo{person}{JD Tygar}.} \bibinfo{year}{2013}\natexlab{}.
\newblock \showarticletitle{Approaches to adversarial drift}. In
  \bibinfo{booktitle}{\emph{Proceedings of the 2013 ACM workshop on Artificial
  intelligence and security}}. \bibinfo{pages}{99--110}.
\newblock


\bibitem[Ke et~al\mbox{.}(2017)]%
        {ke2017lightgbm}
\bibfield{author}{\bibinfo{person}{Guolin Ke}, \bibinfo{person}{Qi Meng},
  \bibinfo{person}{Thomas Finley}, \bibinfo{person}{Taifeng Wang},
  \bibinfo{person}{Wei Chen}, \bibinfo{person}{Weidong Ma},
  \bibinfo{person}{Qiwei Ye}, {and} \bibinfo{person}{Tie-Yan Liu}.}
  \bibinfo{year}{2017}\natexlab{}.
\newblock \showarticletitle{Lightgbm: A highly efficient gradient boosting
  decision tree}.
\newblock \bibinfo{journal}{\emph{Advances in neural information processing
  systems}}  \bibinfo{volume}{30} (\bibinfo{year}{2017}),
  \bibinfo{pages}{3146--3154}.
\newblock


\bibitem[Kolosnjaji et~al\mbox{.}(2018)]%
        {KolosnjajiDBMGE18}
\bibfield{author}{\bibinfo{person}{Bojan Kolosnjaji}, \bibinfo{person}{Ambra
  Demontis}, \bibinfo{person}{Battista Biggio}, \bibinfo{person}{Davide
  Maiorca}, \bibinfo{person}{Giorgio Giacinto}, \bibinfo{person}{Claudia
  Eckert}, {and} \bibinfo{person}{Fabio Roli}.}
  \bibinfo{year}{2018}\natexlab{}.
\newblock \showarticletitle{Adversarial Malware Binaries: Evading Deep Learning
  for Malware Detection in Executables}. In \bibinfo{booktitle}{\emph{The
  European Signal Processing Conference, {EUSIPCO}}}.
  \bibinfo{pages}{533--537}.
\newblock


\bibitem[Li and Li(2017)]%
        {LiL17}
\bibfield{author}{\bibinfo{person}{Xin Li} {and} \bibinfo{person}{Fuxin Li}.}
  \bibinfo{year}{2017}\natexlab{}.
\newblock \showarticletitle{Adversarial Examples Detection in Deep Networks
  with Convolutional Filter Statistics}. In \bibinfo{booktitle}{\emph{{IEEE}
  International Conference on Computer Vision, {ICCV}}}.
  \bibinfo{pages}{5775--5783}.
\newblock


\bibitem[Makandar and Patrot(2017)]%
        {makandar2017malware}
\bibfield{author}{\bibinfo{person}{Aziz Makandar} {and} \bibinfo{person}{Anita
  Patrot}.} \bibinfo{year}{2017}\natexlab{}.
\newblock \showarticletitle{Malware class recognition using image processing
  techniques}. In \bibinfo{booktitle}{\emph{Proceedings of the 2017
  International Conference on Data Management, Analytics and Innovation
  (ICDMAI)}}. \bibinfo{pages}{76--80}.
\newblock


\bibitem[Mercaldo and Santone(2020)]%
        {mercaldo2020deep}
\bibfield{author}{\bibinfo{person}{Francesco Mercaldo} {and}
  \bibinfo{person}{Antonella Santone}.} \bibinfo{year}{2020}\natexlab{}.
\newblock \showarticletitle{Deep learning for image-based mobile malware
  detection}.
\newblock \bibinfo{journal}{\emph{Journal of Computer Virology and Hacking
  Techniques}} (\bibinfo{year}{2020}), \bibinfo{pages}{1--15}.
\newblock


\bibitem[Metzen et~al\mbox{.}(2017)]%
        {MetzenGFB17}
\bibfield{author}{\bibinfo{person}{Jan~Hendrik Metzen}, \bibinfo{person}{Tim
  Genewein}, \bibinfo{person}{Volker Fischer}, {and} \bibinfo{person}{Bastian
  Bischoff}.} \bibinfo{year}{2017}\natexlab{}.
\newblock \showarticletitle{On Detecting Adversarial Perturbations}. In
  \bibinfo{booktitle}{\emph{the 5th International Conference on Learning
  Representations, {ICLR}}}.
\newblock


\bibitem[Miyato et~al\mbox{.}(2016)]%
        {miyatoMKNI15}
\bibfield{author}{\bibinfo{person}{Takeru Miyato}, \bibinfo{person}{Shin-ichi
  Maeda}, \bibinfo{person}{Masanori Koyama}, \bibinfo{person}{Ken Nakae}, {and}
  \bibinfo{person}{Shin Ishii}.} \bibinfo{year}{2016}\natexlab{}.
\newblock \showarticletitle{Distributional smoothing with virtual adversarial
  training}. In \bibinfo{booktitle}{\emph{International Conference on Learning
  Representations.}} \bibinfo{pages}{1--12}.
\newblock


\bibitem[Mohaisen et~al\mbox{.}(2015)]%
        {MohaisenAM15}
\bibfield{author}{\bibinfo{person}{Aziz Mohaisen}, \bibinfo{person}{Omar
  Alrawi}, {and} \bibinfo{person}{Manar Mohaisen}.}
  \bibinfo{year}{2015}\natexlab{}.
\newblock \showarticletitle{{AMAL:} High-fidelity, behavior-based automated
  malware analysis and classification}.
\newblock \bibinfo{journal}{\emph{Computers {\&} Security}}
  \bibinfo{volume}{52} (\bibinfo{year}{2015}), \bibinfo{pages}{251--266}.
\newblock


\bibitem[Moosavi{-}Dezfooli et~al\mbox{.}(2016)]%
        {Moosavi-Dezfooli16}
\bibfield{author}{\bibinfo{person}{Seyed{-}Mohsen Moosavi{-}Dezfooli},
  \bibinfo{person}{Alhussein Fawzi}, {and} \bibinfo{person}{Pascal Frossard}.}
  \bibinfo{year}{2016}\natexlab{}.
\newblock \showarticletitle{{DeepFool}: {A} Simple and Accurate Method to Fool
  Deep Neural Networks}. In \bibinfo{booktitle}{\emph{{IEEE} Conference on
  Computer Vision and Pattern Recognition}}. \bibinfo{pages}{2574--2582}.
\newblock


\bibitem[Nataraj et~al\mbox{.}(2011)]%
        {nataraj2011malware}
\bibfield{author}{\bibinfo{person}{Lakshmanan Nataraj},
  \bibinfo{person}{Sreejith Karthikeyan}, \bibinfo{person}{Gregoire Jacob},
  {and} \bibinfo{person}{BS Manjunath}.} \bibinfo{year}{2011}\natexlab{}.
\newblock \showarticletitle{Malware images: visualization and automatic
  classification}. In \bibinfo{booktitle}{\emph{Proceedings of the 8th
  international symposium on visualization for cyber security}}.
  \bibinfo{pages}{4}.
\newblock


\bibitem[Ni et~al\mbox{.}(2018)]%
        {NiQZ18}
\bibfield{author}{\bibinfo{person}{Sang Ni}, \bibinfo{person}{Quan Qian}, {and}
  \bibinfo{person}{Rui Zhang}.} \bibinfo{year}{2018}\natexlab{}.
\newblock \showarticletitle{Malware identification using visualization images
  and deep learning}.
\newblock \bibinfo{journal}{\emph{Computers {\&} Security}}
  \bibinfo{volume}{77} (\bibinfo{year}{2018}), \bibinfo{pages}{871--885}.
\newblock


\bibitem[Raff et~al\mbox{.}(2018)]%
        {RaffBSBCN18}
\bibfield{author}{\bibinfo{person}{Edward Raff}, \bibinfo{person}{Jon Barker},
  \bibinfo{person}{Jared Sylvester}, \bibinfo{person}{Robert Brandon},
  \bibinfo{person}{Bryan Catanzaro}, {and} \bibinfo{person}{Charles~K.
  Nicholas}.} \bibinfo{year}{2018}\natexlab{}.
\newblock \showarticletitle{Malware Detection by Eating a Whole {EXE}}. In
  \bibinfo{booktitle}{\emph{The Workshops of the The Thirty-Second {AAAI}
  Conference on Artificial Intelligence, New Orleans, Louisiana, USA, February
  2-7, 2018}} \emph{(\bibinfo{series}{{AAAI} Workshops},
  Vol.~\bibinfo{volume}{{WS-18}})}. \bibinfo{publisher}{{AAAI} Press},
  \bibinfo{pages}{268--276}.
\newblock


\bibitem[Raff et~al\mbox{.}(2017)]%
        {RaffSN17}
\bibfield{author}{\bibinfo{person}{Edward Raff}, \bibinfo{person}{Jared
  Sylvester}, {and} \bibinfo{person}{Charles Nicholas}.}
  \bibinfo{year}{2017}\natexlab{}.
\newblock \showarticletitle{Learning the {PE} Header, Malware Detection with
  Minimal Domain Knowledge}. In \bibinfo{booktitle}{\emph{Proceedings of the
  10th {ACM} Workshop on Artificial Intelligence and Security, AISec@CCS
  2017}}. \bibinfo{publisher}{{ACM}}, \bibinfo{pages}{121--132}.
\newblock


\bibitem[Su et~al\mbox{.}(2018)]%
        {SuVPSFS18}
\bibfield{author}{\bibinfo{person}{Jiawei Su},
  \bibinfo{person}{Danilo~Vasconcellos Vargas}, \bibinfo{person}{Sanjiva
  Prasad}, \bibinfo{person}{Daniele Sgandurra}, \bibinfo{person}{Yaokai Feng},
  {and} \bibinfo{person}{Kouichi Sakurai}.} \bibinfo{year}{2018}\natexlab{}.
\newblock \showarticletitle{Lightweight Classification of IoT Malware Based on
  Image Recognition}. In \bibinfo{booktitle}{\emph{{IEEE} Annual Computer
  Software and Applications Conference, {COMPSAC}}}. \bibinfo{publisher}{{IEEE}
  Computer Society}, \bibinfo{pages}{664--669}.
\newblock


\bibitem[Suciu et~al\mbox{.}(2019)]%
        {SuciuCJ19}
\bibfield{author}{\bibinfo{person}{Octavian Suciu}, \bibinfo{person}{Scott~E.
  Coull}, {and} \bibinfo{person}{Jeffrey Johns}.}
  \bibinfo{year}{2019}\natexlab{}.
\newblock \showarticletitle{Exploring Adversarial Examples in Malware
  Detection}. In \bibinfo{booktitle}{\emph{2019 {IEEE} Security and Privacy
  Workshops, {SP} Workshops 2019, San Francisco, CA, USA, May 19-23, 2019}}.
  \bibinfo{publisher}{{IEEE}}, \bibinfo{pages}{8--14}.
\newblock


\bibitem[Vasan et~al\mbox{.}(2020)]%
        {vasan2020image}
\bibfield{author}{\bibinfo{person}{Danish Vasan}, \bibinfo{person}{Mamoun
  Alazab}, \bibinfo{person}{Sobia Wassan}, \bibinfo{person}{Babak Safaei},
  {and} \bibinfo{person}{Qin Zheng}.} \bibinfo{year}{2020}\natexlab{}.
\newblock \showarticletitle{Image-based malware classification using ensemble
  of CNN architectures (IMCEC)}.
\newblock \bibinfo{journal}{\emph{Computers \& Security}}
  (\bibinfo{year}{2020}), \bibinfo{pages}{101748}.
\newblock


\bibitem[Wang et~al\mbox{.}(2017)]%
        {WangGZOXLG17}
\bibfield{author}{\bibinfo{person}{Qinglong Wang}, \bibinfo{person}{Wenbo Guo},
  \bibinfo{person}{Kaixuan Zhang}, \bibinfo{person}{Alexander G.~Ororbia II},
  \bibinfo{person}{Xinyu Xing}, \bibinfo{person}{Xue Liu}, {and}
  \bibinfo{person}{C.~Lee Giles}.} \bibinfo{year}{2017}\natexlab{}.
\newblock \showarticletitle{Adversary Resistant Deep Neural Networks with an
  Application to Malware Detection}. In \bibinfo{booktitle}{\emph{Proceedings
  of the 23rd {ACM} {SIGKDD} International Conference on Knowledge Discovery
  and Data Mining}}. \bibinfo{pages}{1145--1153}.
\newblock


\bibitem[Xu et~al\mbox{.}(2018)]%
        {Xu0Q18}
\bibfield{author}{\bibinfo{person}{Weilin Xu}, \bibinfo{person}{David Evans},
  {and} \bibinfo{person}{Yanjun Qi}.} \bibinfo{year}{2018}\natexlab{}.
\newblock \showarticletitle{Feature Squeezing: Detecting Adversarial Examples
  in Deep Neural Networks}. In \bibinfo{booktitle}{\emph{the Network and
  Distributed System Security Symposium, {NDSS}}}.
\newblock


\bibitem[Yousefi-Azar et~al\mbox{.}(2018)]%
        {yousefi2018malytics}
\bibfield{author}{\bibinfo{person}{Mahmood Yousefi-Azar},
  \bibinfo{person}{Leonard~GC Hamey}, \bibinfo{person}{Vijay Varadharajan},
  {and} \bibinfo{person}{Shiping Chen}.} \bibinfo{year}{2018}\natexlab{}.
\newblock \showarticletitle{Malytics: a malware detection scheme}.
\newblock \bibinfo{journal}{\emph{IEEE Access}}  \bibinfo{volume}{6}
  (\bibinfo{year}{2018}), \bibinfo{pages}{49418--49431}.
\newblock


\bibitem[Zhang et~al\mbox{.}(2016)]%
        {ZhangQYOH16}
\bibfield{author}{\bibinfo{person}{Jixin Zhang}, \bibinfo{person}{Zheng Qin},
  \bibinfo{person}{Hui Yin}, \bibinfo{person}{Lu Ou}, {and}
  \bibinfo{person}{Yupeng Hu}.} \bibinfo{year}{2016}\natexlab{}.
\newblock \showarticletitle{{IRMD:} Malware Variant Detection Using Opcode
  Image Recognition}. In \bibinfo{booktitle}{\emph{Proceedings of the 22nd
  {IEEE} International Conference on Parallel and Distributed Systems,
  {ICPADS}}}. \bibinfo{pages}{1175--1180}.
\newblock


\bibitem[Zhang et~al\mbox{.}(2020)]%
        {ZhangQW20}
\bibfield{author}{\bibinfo{person}{Zhaoqi Zhang}, \bibinfo{person}{Panpan Qi},
  {and} \bibinfo{person}{Wei Wang}.} \bibinfo{year}{2020}\natexlab{}.
\newblock \showarticletitle{Dynamic Malware Analysis with Feature Engineering
  and Feature Learning}. In \bibinfo{booktitle}{\emph{The {AAAI} Conference on
  Artificial Intelligence, {AAAI}}}. \bibinfo{publisher}{{AAAI} Press},
  \bibinfo{pages}{1210--1217}.
\newblock


\end{thebibliography}

\appendix

\begin{table*}[h]
\centering
\caption{The baseline comparison results for malware classification task. PR: Padding Removal, SS: Software Stripping, BR: Bytes Resetting, Mod.: information and bytes values manipulation, C1: the reported accuracy is detecting \textit{Downloader} malware samples, C2: the reported accuracy is detecting \textit{Spyware} malware samples.}
\label{tab:claasification_baselinecomparison}\vspace{-3mm}
\scalebox{0.82}{
\begin{tabular}{!{\vrule width 1.2pt}l!{\vrule width 1.2pt}c|c|c!{\vrule width 1.2pt}c|c|c!{\vrule width 1.2pt}c|c|c!{\vrule width 1.2pt}c|c|c!{\vrule width 1.2pt}c|c|c!{\vrule width 1.2pt}c|c|c!{\vrule width 1.2pt}}
\Xhline{3\arrayrulewidth}
    \multirow{3}{*}{Baseline}  & \multicolumn{3}{c!{\vrule width 1.2pt}}{\multirow{2}{*}{Pre-processing}} & \multicolumn{15}{c!{\vrule width 1.2pt}}{Performance}\\\cline{5-19} 
    & \multicolumn{3}{c!{\vrule width 1.2pt}}{} & \multicolumn{3}{c!{\vrule width 1.2pt}}{Baseline} & \multicolumn{3}{c!{\vrule width 1.2pt}}{Header Mod.} & \multicolumn{3}{c!{\vrule width 1.2pt}}{Inter-section Mod.} & \multicolumn{3}{c!{\vrule width 1.2pt}}{Binary Padding} & \multicolumn{3}{c!{\vrule width 1.2pt}}{Section Injection}\\\cline{2-19} 
& PR  & SS & BR & F-1 & C1 & C2 & F-1 & C1 & C2 & F-1 & C1 & C2 & F-1 & C1 & C2 & F-1 & C1 & C2\\
\Xhline{2\arrayrulewidth}

\multirow{5}{*}{Ember} & \ecir &  \ecir &  \ecir & 88.24 & 87.90 & 88.50 & 87.70 & 88.60 & 87.00 &  88.22 & 87.90 & 88.49      &  35.40 & 54.00 & 31.40     &  83.72 & 77.35 & 88.29   \\\cline{2-19} 
 & \fcir &  \ecir &  \ecir &  87.59 & 86.77 & 88.20 & 87.61 & 87.47 & 87.70 &  87.57 & 86.77 & 88.19    &  87.59 & 86.77 & 88.20     &  87.31 & 86.17 & 88.19  \\\cline{2-19} 
 & \ecir &  \fcir &  \ecir &    87.46 & 86.90 & 87.90 &  87.46 & 86.90 & 87.90  &   87.46 & 86.90 & 87.90     &  23.85 & 23.50 & 23.90     &  83.60 & 78.16 & 87.49 \\\cline{2-19} 
 & \ecir &  \ecir &  \fcir &    87.51 & 86.90 & 87.99  & 87.39 & 87.50 & 87.30 &    87.51 & 86.90 & 87.99      &  33.31 & 64.80 & 27.02     &   83.35 & 77.15 & 87.78   \\\cline{2-19}
 & \fcir &  \fcir &  \fcir &   87.73 & 87.88 & 87.60 &  87.73 & 87.88 & 87.60  & 87.73 & 87.88 & 87.60      &   87.73 & 87.88 & 87.60      &  84.49 & 80.36 & 87.49  \\\Xhline{3\arrayrulewidth}

\multirow{5}{*}{Ember (Mono)} & \ecir &  \ecir &  \ecir & 87.82 & 86.70 & 88.70 & 87.33 & 87.10 & 87.50 & 87.81 & 86.70 & 88.69      & 75.30 & 68.00 & 79.70     &    76.44 & 78.96 & 74.87 \\\cline{2-19} 
 & \fcir &  \ecir &  \ecir & 87.51 & 86.47 & 88.30 & 87.67 & 86.97 & 88.20  &  87.50 & 86.47 & 88.29     &  87.51 & 86.47 & 88.30     &  79.04 & 83.87 & 75.88   \\\cline{2-19} 
 & \ecir &  \fcir &  \ecir &   87.92 & 86.80 & 88.80 &  87.92 & 86.80 & 88.80  & 87.92 & 86.80 & 88.80  &   68.94 & 38.60 & 84.90  &  80.00 & 83.47 & 77.68  \\\cline{2-19} 
 & \ecir &  \ecir &  \fcir & 87.42 & 86.70 & 87.99 & 87.31 & 87.20 & 87.40 &  87.42 & 86.70 & 87.99    &  75.68 & 69.30 & 79.58     &  76.51 & 79.36 & 74.75  \\\cline{2-19} 
 & \fcir &  \fcir &  \fcir & 87.06 & 86.07 & 87.80  & 87.06 & 86.07 & 87.80 &  87.06 & 86.07 & 87.80     & 87.06 & 86.07 & 87.80    &   80.08 & 83.67 & 77.68  \\\Xhline{3\arrayrulewidth}

 \multirow{5}{*}{Software Visualization} & \ecir &  \ecir &  \ecir & 85.13 & 82.60 & 87.00  & 83.73 & 81.00 & 85.70 &   85.11 & 82.70 & 86.90      &  65.29 & 46.50 & 74.40     &  79.83 & 70.04 & 86.30 \\\cline{2-19} 
 & \fcir &  \ecir &  \ecir & 84.88 & 81.96 & 87.00 & 83.61 & 80.66 & 85.70  & 84.90 & 82.16 & 86.90    &  84.88 & 81.96 & 87.00    &  84.70 & 81.26 & 87.20 \\\cline{2-19} 
 & \ecir &  \fcir &  \ecir &  84.52 & 82.50 & 86.00  &  84.52 & 82.50 & 86.00 &  84.56 & 82.60 & 86.00    &  66.64 & 37.10 & 81.39    &   81.65 & 73.85 & 87.00  \\\cline{2-19} 
 & \ecir &  \ecir &  \fcir & 83.97 & 82.00 & 85.39 &  82.23 & 80.7 & 83.39 &  83.97 & 82.0 & 85.39   & 65.31 & 44.90 & 75.20    &   79.11 & 69.64 & 85.29 \\\cline{2-19}
 & \fcir &  \fcir &  \fcir &  84.17 & 81.76 & 85.90 &  84.17 & 81.76 & 85.90  &  84.17 & 81.76 & 85.90   &   84.17 & 81.76 & 85.90      &    81.50 & 73.15 & 87.20  \\\Xhline{3\arrayrulewidth}

\multirow{5}{*}{MalConv} & \ecir &  \ecir &  \ecir & 84.79 & 85.60 & 84.20 &  82.12 & 71.10 & 89.80 &  84.87 & 85.60 & 84.44     &  83.95 & 83.60 & 84.20     &  84.73 & 84.75 & 84.60 \\\cline{2-19} 
 & \fcir &  \ecir &  \ecir &  85.38 & 85.76 & 85.00 & 82.97 & 72.22 & 90.40 & 85.35 & 85.76 & 85.05    &  85.38 & 85.76 & 85.00     &    85.43 & 85.45 & 85.30 \\\cline{2-19} 
 & \ecir &  \fcir &  \ecir & 85.98 & 85.30 & 86.50 &  85.98 & 85.30 & 86.50  &   85.98 & 85.30 & 86.50     & 85.21 & 83.60 & 86.40     &  85.99 & 82.63 & 88.40  \\\cline{2-19} 
 & \ecir &  \ecir &  \fcir & 84.47 & 85.90 & 83.54 & 79.18 & 65.10 & 88.40  &  84.47 & 85.90 & 83.54    &  83.97 & 84.30 & 83.84     &   83.74 & 83.43 & 84.08  \\\cline{2-19} 
 & \fcir &  \fcir &  \fcir &  84.05 & 82.65 & 85.00  &  84.05 & 82.65 & 85.00      &  84.05 & 82.65 & 85.00   &   84.05 & 82.65 & 85.00   &  85.18 & 82.55 & 87.10 \\\Xhline{3\arrayrulewidth}

\multirow{5}{*}{Graph Encoding (Ours)}  & \ecir &  \ecir &  \ecir & 87.61 & 87.03 & 88.19 & 86.62 & 86.58 & 86.80  & 87.54 & 87.04 & 88.08    & 87.11 & 86.45 & 87.78  &   86.29 & 85.09 & 87.36\\\cline{2-19} 
 & \fcir &  \ecir &  \ecir &  87.41 & 85.93 & 88.71 & 86.83 & 85.79 & 87.81  &   87.75 & 86.75 & 88.69     &  87.41 & 85.93 & 88.71   &   85.64 & 87.25 & 84.58  \\\cline{2-19} 
 & \ecir &  \fcir &  \ecir & 87.01 & 85.77 & 88.13 &  87.54 & 86.58 & 88.43   &   86.74 & 86.09 & 87.41   & 86.6 & 85.92 & 87.31  &   84.79 & 86.88 & 83.40 \\\cline{2-19} 
 & \ecir &  \ecir &  \fcir & 87.25 & 85.73 & 88.59 &  86.12 & 85.99 & 86.39  &   86.87 & 85.53 & 88.08     &  87.43 & 86.55 & 88.28  &  85.43 & 85.73 & 85.39 \\\cline{2-19} 
 & \fcir &  \fcir &  \fcir & 87.20 & 85.60 & 88.63 &  87.20 & 85.60 & 88.63  & 87.20 & 85.60 & 88.63     & 87.20 & 85.60 & 88.63   &   85.92 & 88.42 & 84.21  \\

\Xhline{3\arrayrulewidth}

\end{tabular}}\vspace{-1mm}

\end{table*}

\section{Malware Classification Capabilities}\label{sec:classification_task_analysis}
Malware classification task is crucial to ensure that our proposed modifications and detection scheme preserve the various capabilities and patterns of different malware families. 
To this end, we compiled a dataset of 2,000 \textit{``Downloader''} malicious samples and \textit{``Spyware''} malicious samples.
The evaluation results are shown in~\autoref{tab:claasification_baselinecomparison}. In this process, we restricted the training process to malware samples, not incorporating any benign samples.

\BfPara{Baseline Performance Comparison} 
The EMBER-based model provides state-of-the-art classification performance, with an F-1 score of 88.24\%. Notice that the performance reduction upon software pre-processing is marginal, indicating that the pre-processing approaches preserved the malicious capabilities.

\noindent\textbf{\underline{\textit{Key Takeaway:}}} Removing volatile information channels through software pre-processing did not affect the malware classification capabilities, preserving the patterns and allowing robust capturing of them.

\BfPara{Header Information Stripping} 
As shown in table~\ref{tab:claasification_baselinecomparison}, the header information manipulation attack only affected MalConv detection accuracy, with marginal effects on the remaining detection approaches. That being said, similar to the detection task, software stripping pre-processing, the attack channel is eliminated, and the models report exact performance on both baseline and header information modified binaries.

\noindent\textbf{\underline{\textit{Key Takeaway:}}} While header information manipulation attack is not as effective in classification settings, software stripping pre-processing is still capable of eliminating the header manipulation attack channel, invalidating its effects on the baseline classification accuracy.

\BfPara{Inter-section Injection} Acknowledging the limited capabilities of this attack, we proceed with eliminating the attack channel using the inter-section bytes setting pre-processing method. That being said, this attack did not show any effectiveness in successfully misclassifying malware samples into a different family. 

\noindent\textbf{\underline{\textit{Key Takeaway:}}} The observations of the classification task are in line with the previous task, where inter-section bytes resetting effectively eliminates the effects of inter-section injection, including potential code caves and targeted adversarial binaries.

\BfPara{Binaries Padding} 
Similar to the detection task, we observed a significant reduction of classification accuracy upon padding binaries at the end of the malicious binaries. However, this decrease is not applicable to our proposed graph encoding model. This is attributed to the fact that our proposed approach is per-component (i.e., section) limited, and is not affected by out-of-boundary binary padding.

\noindent\textbf{\underline{\textit{Key Takeaway:}}} Binary padding is effective under the classification task, significantly reducing the classification performance of various models and engines. However, using padding removal pre-processing eliminates its adversarial effects.

\BfPara{Section Injection} 
In this attack, we are injecting random malicious sections of \textit{Spyware Malware} into \textit{Downloader Malware}, and vice versa. Notice that the performance of all the models was reduced. However, MalConv and the proposed graph encoding baseline model performance reduction is minimal, with a higher F-1 score associated with the proposed approach, resulting in state-of-the-art classification performance.

\noindent\textbf{\underline{\textit{Key Takeaway:}}} The proposed per-section graph representation is capable of holding the best classification performance even after the injection of the malicious sections from another malware family. Further, in line with the detection task observation, the proposed pre-processing techniques and modifications are incapable of mitigating the effects of this attack.

\end{document}